\documentclass[11pt,onecolumn,journal]{article}

\usepackage[bottom=2.5cm,top=2.5cm,left=2.5cm,right=2.5cm]{geometry}

\usepackage{pdfpages}
\usepackage{pgfplots}
\usepackage{subfigure}
\usetikzlibrary{spy,calc}
\usepackage{CJK}
\usepackage{multirow}
\usepackage{pifont}
\usepackage{epsfig}
\usepackage{array}
\usepackage[toc]{appendix}

\usepackage{amsmath,amssymb,amsthm,graphicx,caption,cases}	
\usepackage[bottom]{footmisc}	
\usepackage{algorithmic}

\usepackage{abstract,amsfonts,amsbsy,amstext,mathrsfs,latexsym,color,url,xcolor,authblk,fontenc,bm}

\definecolor{NUSBlue}{RGB}{0,61,124} 
\definecolor{NUSOrange}{RGB}{239,124,0}
\usepackage[colorlinks=false,allbordercolors=NUSOrange]{hyperref}

\DeclareOldFontCommand{\bf}{\normalfont\bfseries}{\mathbf}

\DeclareMathOperator{\diag}{diag}

\newcommand{\bV}{\mathbf{V}}
\newcommand{\bI}{\mathbf{I}}

\newcommand{\bD}{\mathbf{D}}
\newcommand{\bP}{\mathbf{P}}

\newcommand{\bU}{\mathbf{U}}

\newcommand{\bW}{\mathbf{W}}
\newcommand{\bX}{\mathbf{X}}
\newcommand{\bY}{\mathbf{Y}}

\newcommand{\bz}{\mathbf{z}}

\newcommand{\bv}{\mathbf{v}}

\newcommand{\bb}{\mathbf{b}}
\newcommand{\bc}{\mathbf{c}}
\newcommand{\bd}{\mathbf{d}}

\newcommand{\bu}{\mathbf{u}}
\newcommand{\bs}{\mathbf{s}}
\newcommand{\bt}{\mathbf{t}}
\newcommand{\bA}{\mathbf{A}}

\newcommand{\bx}{\mathbf{x}}
\newcommand{\by}{\mathbf{y}}

\newcommand{\R}{\mathbb R}

\definecolor{NUSBlue}{RGB}{0,61,124}   

\usepackage{tikz}
\usepackage{mathtools}

\def\nudge{.5}

\tikzset{axis/.style={ultra thin, Grey, -latex, shorten <=-\nudge cm, shorten >=-2*\nudge cm}}
\tikzset{line/.style={thick}}

\usepackage{textcomp}
\DeclareMathAlphabet\mathbfcal{OMS}{cmsy}{b}{n}
\usepackage[linesnumbered,vlined,ruled]{algorithm2e}
\theoremstyle{plain}
\newtheorem{thm}{Theorem}

\newtheoremstyle{cited}%
  {3pt}
  {3pt}
{\itshape}
  {}
  {\bfseries}
  {.}
  {.5em}
  {\thmname{#1} \thmnumber{#2} \thmnote{\normalfont#3}}
\theoremstyle{cited}
\newtheorem{citedthm}[thm]{Theorem}
\newtheorem{citedlem}[thm]{Lemma}
\newtheorem{citedcor}[thm]{Corollary}

\newtheorem{citedprop}[thm]{Proposition}

\usepackage{graphicx,epstopdf,epsf} 
\usepackage{cite}
\usepackage{tabularx}
\usepackage{array}
\usepackage{booktabs}
\usepackage{comment}

\begin{document}
\title{Learning DNN networks using un-rectifying ReLU with compressed sensing application}\date{\today}
\author{Wen-Liang Hwang and Shih-Shuo Tung}
\maketitle

\begin{abstract}
The success of deep neural networks (DNNs) depends largely on the means by which training data is used to learn the network. Most learning algorithms are based on gradient descent and block coordinate descent approaches. Under these schemes, a non-linear activation function is directly applied throughout the course of optimization, which means that only the input and output of the activation function are optimized. By contrast, the ``un-rectifying" technique expresses a non-linear point-wise (non-smooth) activation function as a data-dependent variable (activation variable), which means that the activation variable along with its input and output can all be employed in optimization. The fact that any continuous point-wise piecewise linear activation function can be expressed as a DNN composed entirely of ReLUs makes it possible to examine the process of learning networks composed entirely of ReLU activations. 
The fact that the ReLU network in this study was un-rectified means that the activation functions could be  replaced with data-dependent activation variables in the form of equations and constraints. The discrete nature of activation variables associated with un-rectifying ReLUs allows the reformulation of deep learning problems as problems of combinatorial optimization. However, we demonstrate that the optimal solution to a combinatorial optimization problem can be preserved by relaxing the discrete domains of activation variables to closed intervals. This makes it easier to learn a network using methods developed for real-domain constrained optimization. We also demonstrate that by introducing data-dependent slack variables as constraints, it is possible to optimize a network based on the augmented Lagrangian approach. This means that our method could theoretically achieve global convergence and all limit points are critical points of the learning problem. In experiments, our novel approach to solving the compressed sensing recovery problem achieved state-of-the-art performance when applied to the MNIST database and natural images.

\end{abstract}

\section{Introduction}

Deep neural networks (DNNs) are an indispensable tool for data-driven approaches to problem-solving. They have demonstrated outstanding performance in computer vision, speech recognition, gaming, and signal/image processing. Current research into DNNs focuses novel techniques from the perspective of 
comprehensiveness\cite{balestriero2018mad,Wen19}, structure compactness\cite{gong2014compressing}, classification performance\cite{he2016deep}, and effiency\cite{gregor2010learning} (e.g., converting an inverse problem into a forward inference problem). Examples include but are not limited to constructing an explainable non-linear system with classification performance comparable to a DNN~\cite{kuo2018data},
simplifying the complexity of DNNs without compromising performance via pruning followed by quantization\cite{szegedy2016inception,kung2020term}, improving performance by increasing the depths/and/or connections\cite{huang2017densely}, and using the unrolling technique\cite{zhang2018ista}. The success of DNNs depends heavily on the means by which training data is used to train the network. 
The objective in DNN learning is to identify a manipulation capable of converting model input into the model output,
tailored to a particular problem in a specific domain, through the application of regularization terms to weight matrices, activation functions, and a priori knowledge of input/output data. 
In addition to the popular back-propagation algorithm and its variants, block coordinate descent (BCD) and the alternating direction method of multipliers (ADMM) are alternative approaches to learning a DNN which avoid the problem of gradient instability. The BCD approach involves decomposing a problem into sub-problems to be solved cyclically, while the remaining blocks remain fixed at their last updated values. The ADMM approach searches for saddle points of augmented Lagrangian functions via primal and dual variable updates, wherein the updating of primal variables is usually based on the BCD approach.

Nearly all approaches to DNN learning are based on the equation $a = \rho z$, where $\rho$ is an activation function. For example, the function of the equation can be formualted specifically for minimization, as follows: $\| a  - \rho z\|^2$\cite{zeng2018global,gao2020admm}. Most theoretical studies on the convergence of DNN learning make assumptions pertaining to smoothness. If $\rho$ is not a smooth activation function, then it must be approximated. The ``un-rectifying" technique \cite{Wen19} expresses a non-linear point-wise (non-smooth) activation function as a data-dependent variable. For example, $a = \rho z$ becomes $a = d z$, where $d$ is a data-dependent variable (i.e., its value is dependent on the value of $z$). While $\rho$ is not involved directly in optimization, $d$ can be used in optimization in the same way as other variables. It is from this perspective that we explore DNN learning in this paper.

Any continuous point-wise piecewise linear activation function can be expressed as a DNN composed entirely of ReLU activation functions~\cite{Ara18}. Furthermore, a ReLU activation function is a data-dependent operator, used to partition the input space into polytopes\cite{Wen19}. Thus, a DNN with continuous point-wise piecewise linear activation functions can be viewed as a piecewise affine linear function; i.e., affine linear with regard to the polytopes into which the input space is partitioned. The explicit expression of domains and affine mappings can be characterized using the un-rectifying technique \cite{Wen19}, in which continuous point-wise piecewise linear activation functions are replaced with data-dependent equations, each of which comprises a diagonal matrix with entries $\{0, 1\}$ and constraints. When the activation functions of a network are un-rectified layer-by-layer, the network is rendered as a composition of data-dependent equations over constraints. This makes it possible to treat learning problems related to a non-linear network as data-dependent constrained optimization problems. 
However, deriving a solution via un-rectified DNN is not a straightforward task, due to the fact that different inputs tend to be associated with different optimization objectives and constraints. Fortunately, although the input is a continuous value in the input space, the number of data-dependent sub-problems is finite, with an upper bound determined by the exponential number of domain polytopes resulting from the combination of values derived from discrete activation functions\cite{serra2017bounding}. In this study, we investigated the issue of convergence in learning an L-layer ReLU network using the un-rectifying approach to the augmented Lagrangian method, in which the ReLU network is represented by the following:
\begin{align*}
\mathcal M_L  = M_L \rho M_{L-1} \ldots \rho M_1,
\end{align*}
where $\rho$ refers to ReLU and $M_i$ is an affine transform, the linear portion of which is $\bW_i$ and the bias is $\bb_i$\cite{HintonReLU,GlorotBordesBengio2011}. This model is simple and restricted but sufficiently general in its ability to approximate any function in $L^p(R^n)$ with $1 \leq p \leq \infty$ as long as the number of ReLUs/hidden layers is not a concern\cite{Ara18,heinecke2020refinement}. Furthermore, any point-wise piecewise linear activation function can be expressed using a ReLU DNN\footnote{For example, 
the max-pooling of $x_1$ and $x_2$ is defined as
\begin{align*}
&\max(x_1, x_2)  = \frac{x_1 + x_2}{2} + \frac{|x_1 - x_2|}{2} \\
=& \frac{1}{2} (\rho(x_1 + x_2) - \rho(-(x_1 + x_2))+ \frac{1}{2} (\rho (x_1-x_2) + \rho(x_2 - x_1)).
\end{align*}
}.
One daunting obstacle to learning an un-rectified ReLU network lies in optimizing combinatorial problems, due to the fact that the activation variables of ReLUs are in discrete domains following the un-rectifying process.

Optimization over activation variables (equal to the number of ReLUs in a network) of $\{0, 1\}$ is inefficient and manipulating variables in discrete domain means that the optimization is not continuous. A slight change to the input of a variable can generate a discontinuous jump in the value of the variable. Thus, we relax the discrete domains of activation variables to closed intervals to allow optimization of the combinatorial problem in continuous real domains. The validity of this approach is demonstrated by the fact that the optimal solution to the original learning problem is equivalent to the optimal solution to the relaxed learning problem. Moreover, the fact that the constraints imposed by the relaxation of un-rectifying ReLUs can be expressed as twice-continuously-differentiable functions with continuous (activation) variables comprising simple bounds, rendering the learning problem computationally tractable. The problem can therefore be solved efficiently using the constrained optimization approach. 
Furthermore, after introducing data-dependent slack variables (converting inequality constraints into equality constraints), the un-rectified network can be optimally trained within the framework of the Conn, Gould, and Toint (CGT)-algorithm\cite{conn1991globally}.
The CGT-algorithm is a global convergence algorithm\footnote{Global convergence means that the iterators generated by the algorithm from any initial point converge to limit points for which necessary conditions of optimality hold.}, which uses the augmented Lagrangian approach to solve problems involving equality constraints and variables with simple bounds, in situations where the objective and all of the constraints are twice-continuously-differentiable. 
Conn, Gould, and Toint considered the problem of finding a local minimizer for the following function:
\begin{align} \label{P1a}
f(\bx^1,\bx^2) \colon \R^n \to \R, \text{ where $\bx^1 \in \R^{n_1}$ and $\bx^2 \in\R^{n_2}$}
\end{align}
where $n = n_1 + n_2$ and $(\bx^1, \bx^2)$ must satisfy 
\begin{align} \label{P2a}
c_i(\bx^1, \bx^2) = 0,  \text{ $1 \leq i \leq m$}.
\end{align}
There is no constraint on $\bx^1$ and any component $x_j$ of $\bx^2$ must satisfy the following simple bounds 
\begin{align} \label{P3a}
l_j \leq x_j \leq u_j
\end{align}
where $u_j > l_j$, $l_j \in \R$, and $u_j \in \R \cup \{\infty\}$. Here, $B$ is used to denote the convex set of $\bx = (\bx^1, \bx^2)$ wherein $\bx^2$ satisfies (\ref{P3a}). 

The main contributions of this paper are as follows:
 \begin{itemize}
 \item We explore the issue of DNN learning in which activation functions can be expressed as data-dependent activation variables in equations and constraints; thereby allowing for the involvement of the  activation variables in optimizations.  In experiments, the proposed learning algorithm achieved state-of-the-art performance for the problem of compressed sensing recovery when applied to the MNIST dataset and natural images. The baseline algorithms used for comparsion were the Adam optimizer\cite{kingma2014adam} (a state-of-the-art implementation of the back-propagation method) and the global convergence block coordinate descent algorithm in \cite{zeng2018global}.
\end{itemize}
Our claims also benefit from a number of technical considerations:
\begin{itemize}
\item We adopted the un-rectifying approach to represent the ReLU network learning problem as a constrained optimization problem with discrete domains of activation variables. We relaxed the discrete entries in activation variables to real domain with values within simple bounded closed intervals. Moreover, we show that the optimal solution to the original problem is equivalent to the optimal solution to the relaxed problem. This transforms a combinatorial problem into a real-domain problem, thereby making it possible to solve the problem of un-rectifying network learning using constrained optimization.
\item We demonstrate that the network learning problem can achieve global convergence to critical points via the augmented Lagrangian approach. This is achieved by weakening constraints imposed on all primal variables with simple bounds\cite{conn1991globally}. Furthermore, the values of the bounds can be any real numbers\footnote{\cite{conn1991globally} gives a detailed derivation when $\bx = \bx^2$ and any variable is bounded with $l_j = 0$ and $u_j = \infty$, whereas this paper gives derivations when $\bx = (\bx^1, \bx^2$) and the values  of $l_j$ and $u_j$ of any component $x_j $ in $\bx^2$ are set to any real numbers.}. The primal variables involving weight matrices, biases, and (simple bounded) activation variables can be updated in accordance with alternating optimization using the three-splitting approach~\cite{taylor2016training,lau2018proximal,zeng2018global}. Empirically, the proposed algorithm is robust to weight initialization and parameter settings. In principle, the proposed algorithm can be customized to learn non-linear networks in which the activation functions are continuous point-wise piecewise linear functions. 
\end{itemize}

The remainder of the paper is organized as follows. Section~\ref{sec:rw} briefly reviews related works, the framework of the CGT-algorithm, and the un-rectifying technique. 
Section~\ref{relaxation} outlines the means by which the un-rectifying and relaxation technique transforms ReLU activation functions into data-dependent equations and equality constraints wherein variables are in continuous domains.
Section~\ref{globalconvergence} presents our theoretical results in which global convergence (using the augmented Lagrangian approach in which all primal variables are updated through alternating minimization) is achieved by weakening the constraints outlined in \cite{conn1991globally}. 
In Section~\ref{representationlearning}, we present the un-rectifying representation of a ReLU network and derive solutions to learn the representation. 
Section~\ref{exp} outlines experiments on compressed sensing demonstrating the numerical convergence of the proposed algorithm, as well as comparisons with other recovery methods. Concluding remarks are presented in Section~\ref{conclusions}.

\section{Related Works} \label{sec:rw}

\subsection{Deep learning algorithms}

Our review of deep learning algorithms deals with the gradient descent approach, alternating minimization (block coordinate descent), and the alternating direction method of multipliers (ADMM), as these are most widely used methods with theoretical support. 
The back-propagation algorithm~\cite{rumelhart1986learning} and its variants are the most popular deep learning algorithms, due to its adaptability to all types of feed-forward neural networks. The fact that back-propagation algorithms are based on the gradient descent approach theoretically guarantees convergence using steps of fixed step size, as long as the objective and activation functions are continuously differentiable in terms of bounded Lipschitz continuous gradients. However, this approach is susceptible to gradient instability and most of the methods designed to deal with this issue necessitate setting an appropriate range of values for the updating of parameters (particularly the learning rate). Fortunately, the problem of  gradient instability has been overcome using weight initialization\cite{glorot10,he2015delving,krizhevsky2017imagenet}, batch-normalization \cite{ioffe2015batch}, and an adaptive learning rate and momentum for the updating of parameters\cite{nesterov1983method,duchi2011adaptive,kingma2014adam}. 
Other techniques that counter gradient instability are reviewed in \cite{ruder2016overview}.
In practice, the large dataset used for training is usually divided into small groups referred to as mini-batches. Stochastic gradient descent treats the sequence of mini-batch training inputs as on-line samples from a specific stochastic process. Many stochastic gradient descent algorithms are variants of the method proposed by Robbbins and Monro~\cite{robbins1951stochastic}, which stipulates the conditions under which the learning rate can be updated to ensure the successive convergence of on-line data to a critical point of the objective function\footnote{The updating of variables at each on-line input can result in a high degree of variation in the value of the objective function. The variance can be pragmatically reduced by performing updates at each mini-batch input.}. The conditions pertaining to convergence at mini-batch inputs were investigated in~\cite{bottou2018optimization}.

The deep learning problem can be formulated in a manner that allows the application of alternating optimization whenever a single block of variables is minimized, while the other variables remain fixed. An outstanding review of the two-splitting and three-splitting approaches for DNN learning can be found in~\cite{zeng2018global}. The fundamental principle underlying two-splitting and three-splitting is the introduction of auxiliary variables to separate highly-coupled DNN structures into separate sub-components and equality constraints. The penalty method is then used to transform the constrained problem into an un-constrained problem, in which equality constraints are replaced as penalty terms in objective function. Below, we illustrate this approach using a three-layer DNN $\mathcal M_3 = M_3 \rho M_2 \rho M_1$ with the following two-splitting formulation~\cite{zhang2017convergent,gu2018fenchel,lau2018proximal} where $\bY$ and $\bX$ are respectively the arrays of $\by_i$ (the training outputs) and $\bx_i$ (the training inputs):
\begin{align} \label{twosplit}
\begin{cases}
\min_{\bV_1, \bV_2, M_1, M_2, M_3} \ell(\bY, M_3 \bV_2) + S(\bV_1,\bV_2) + T(M_1, M_2, M_3) \\
\bV_2 = \rho M_2 \bV_1 \\
\bV_1 = \rho M_1 \bX, 
\end{cases}
\end{align}
where $\ell$ is the loss function, $\rho$ is the non-linear activation function, and $S$ and $T$ are regularization functions. The equality constraint is approximated using the penalty method with penalty parameter $\gamma$ where the constrained problem is transformed into the following unconstrained problem:
\begin{align} \label{unconstrainedtwosplit}
\min_{\bV_1, M_1, M_2} & \ell(\bY, M_3 \bV_2) + S(\bV_1, \bV_2) + T(M_1, M_2, M_3)  \nonumber \\
&+ \gamma (\|\bV_1 - \rho M_1\bX\|_F^2 + \|\bV_2 - \rho M_2\bV_1\|_F^2)
\end{align}
The solution to (\ref{unconstrainedtwosplit}) can be derived using the alternating minimization approach.
In theory, the value of $\gamma$ must gradually increase with an increase in the number of alternating minimization iterations (\ref{unconstrainedtwosplit}), thereby improving approximation to the equality constraint.  In practice however, the value of $\gamma$ is usually a constant set at an appropriate value.
The three-splitting formulation of the three-layer DNN $\mathcal M_3$ is formulated as follows:
\begin{align} \label{threesplit}
\begin{cases}
\min_{\bU_1, \bV_1, M_1, M_2}\ell(\bY, M_3 \bV_2) + S(\bV_1, \bV_2, \bU_1, \bU_2) + T(M_1, M_2, M_3) \\
\bV_1 = \rho \bU_1, 
\bV_2 = \rho \bU_2 \\
\bU_1 = M_1 \bX, \text{ and } \bU_2 = M_2 \bV_1.
\end{cases}
\end{align}
The constraints are dealt with using the penalty method, as in  (\ref{twosplit}).
The two-splitting and three-splitting formulations of the three-layer learning problem are easily extended to generic L-layer DNNs~\cite{taylor2016training}. For alternating minimization with penalty parameters of fixed values, convergence depends on the assumptions related to the loss function (cost function $\ell$), regularization functions (regulators ($S, T$)), and the activation function (non-linearities $\rho$). Recent theoretical results~\cite{zeng2018global} have revealed that the sequence derived by solving the two-splitting/three-splitting formulation would converge to a critical point of the L-layer generalization of (\ref{unconstrainedtwosplit}), as long as the activation functions satisfy Lipschitz continuity over any bounded sets, the loss and regularization functions satisfy the common assumptions in DNN learning. This analysis is based on the K{\L} inequality \cite{kurdyka1998gradients,lojasiewicz1993geometrie} and is strongly influenced by the work of Attouch and Bolte~\cite{attouch2013convergence} and Xu and Yin~\cite{xu2017globally}. As reported in~\cite{zeng2018global}, a DNN can be learned by solving the two-splitting/three-splitting formulation in order to avoid the problem of gradient instability. Despite the fact that this simple scheme leads to global convergence, the solutions to two-splitting and three-splitting formulations tend to be unstable, due to the instability inherent in the penalty method when the values of penalty parameters are large~\cite{bertsekas2014constrained}.  The conditions required  to achieve convergence when using a mini-batch learning algorithm based on the alternating optimization approach are presented in~\cite{choromanska2018beyond}.

The ADMM was originally proposed for convex optimization\cite{glowinski1975approximation}. It has attracted renewed attention due to its applicability to various machine learning and image processing problems. This approach is easily implemented and has proven effective in optimizing sums of fairly simple but non-smooth convex functions as long as high accuracy is not a strict requirement\cite{eckstein2015understanding}. The convergence of non-convex ADMM under a 2-block linear constrain is outlined in \cite{magnusson2015convergence}. The convergences of ADMM under multi-block linear and multi-affine constraints are presented in  \cite{wang2019global} and \cite{gao2020admm}, respectively.  The DNN problem can be formulated/approximated as a non-convex optimization problem with multi-affine constraints. When implemented with two-splitting or three-spilitting formulations, it was demonstrated in \cite{gao2020admm} that the ADMM can be used to solve the DNN learning problem where the primal variables are updated using the block coordinate descent method. However, adhering to the assumptions of convergence in \cite{gao2020admm} requires approximation of non-linear activation using a smoothing technique. Furthermore, a local linear approximation must be applied to obtain a special block pertaining to the properties of assumptions\cite{zeng2019convergence}. 
Global convergence to critical points of DNN learning with sigmoid activation functions was demonstrated in \cite{zeng2020admm}.

\subsection{Augmented Lagrangian approach with simple bounds} \label{sec:globalconv1}

The Lagrangian function and augmented Lagrangian function of (\ref{P1a})-(\ref{P3a}) with respect to constraints $\{c_i\}$ are respectively written as follows:
\begin{align} \label{lagrangian}
L(\bx, \lambda) = f(\bx) + \sum_{i=1}^m \lambda_i c_i(\bx),
\end{align}
and 
\begin{align} \label{lagrangian1}
L_{\mu}(\bx, \lambda) = f(\bx) + \sum_{i=1}^m \lambda_i c_i(\bx) + \frac{1}{2\mu} \sum_{i=1}^m c_i(\bx)^2,
\end{align}
where $\{ \lambda_i\}$ are the Lagrangian multipliers.

Let $\bA(\bx)$ denote the $m \times n$ Jacobian of $c(\bx) = [c_1(\bx), \cdots, c_m(\bx)]^\top$ and let $\bA(\bx)|_J$ denote $m \times |J|$ sub-matrix of $\bA(\bx)$ with columns corresponding to variables (with indices) in $J$. 
Note that the simple bounds (\ref{P3a}) are not included in the augmented Lagrangian function. Rather, it is assumed that sequential minimization used to update the primal variables in the augmented Lagrangian approach ensures that these constraints are always satisfied. 
Global convergence of the CGT-algorithm is derived for the problem
in which the following three conditions are assumed: \\
(AS1) Objective $f(\bx)$ and constraints $\{c_i\}_i$ are twice-continuously-differentiable for $\bx \in B$. \\
(AS2) The considered iterators $\{\bx^{(k)}\}$ lie within a closed, bounded domain, $\Omega \subset B$. \\
(AS3) $J$ denotes the variables (with indices) in $\bx^1$ and the variables (with indices) in $\bx^2$, such that $i \in J$ implies that variable $x_i$ is in $\bx^1$ or $\bx^2$ with a value strictly in the interior of closed interval $[l_i, u_i]$. The column rank of $\bA(\bx^*)|_J$ is no smaller than $m$ at any limit point $\bx^*$ of sequence $\{\bx^{(k)}\}$.\\

The augmented Lagrangian multiplier is generally executed using a bounded penalty parameter for problems involving convex objective functions and constraints.  Nonetheless, the non-convex example in \cite{wang2019global} demonstrates that the bounded penalty parameter is insufficient for the convergence of the augmented Lagrangian multipliers. The CGT-algorithm derives optimal solutions for a class of problems involving smooth objective functions and constraints, which can be non-convex.
The CGT-algorithm is a sophisticated algorithm involving a sequence of non-decreasing and unbounded penalty parameters.  The parameters in the algorithm are carefully updated when the algorithm is executed to ensure that all limit points of convergent sub-sequences are critical points, which meet the variational inequality of the constrained optimization problem. Additionally, the algorithm deals with Lagrangian multiplier estimates $\lambda^{(k)} =[\lambda_1^{(k)}, \cdots, \lambda_m^{(k)}]^\top$ in a manner that 
if $\mu^{(k)}$ converges to zero with an increase in $k$ (corresponding to the unbounding of penalty parameter $\frac{1}{\mu^{(k)}}$ in (\ref{lagrangian1})) when the algorithm is executed, then the product $\mu^{(k)} \|\lambda^{(k)} \|$ converges to zero. 

\subsection{Un-rectifying point-wise piecewise linear activation functions}

The un-rectifying procedure exploits the fact that continuous point-wise piecewise linear activation functions partition the input space into finite number of regions by replacing an activation function with a finite number of data-dependent equations and constraints. Demonstrated in \cite{Wen19} are diagonal matrices and hyper-plane constraints where continuous point-wise piecewise linear activation functions, ReLU and MaxLU (maxpooling following ReLU), are un-rectified. Applying un-rectifying to a ReLU/MaxLU network reveals at least three interesting network properties. First, the input space of such a network is partitioned into polytopes, and adding a layer to a network is equivalent to refining each partitioning region of the network in a way that resembles tree refinement. Second, the stability measured against small input perturbations in a very deep network is linked to the distribution of sparse/compressed weight coefficient in the network. The third issue and main topic in this paper is the fact tht when a network is un-rectified layer-by-layer, it can be expressed as a composition of data-dependent equations and constraints. This allows us to revisit network learning problems from the perspective of constrained optimization.

\section{Un-rectifying ReLUs and ReLU relaxation} \label{relaxation}

Under the un-rectifying approach, the ReLU function becomes a data-dependent variable, which can be analyzed in a manner as other variables.  The ReLU activation function used to map $\R$ to $\R_+$ is not used directly. Rather, it is represented through the introduction of activation variable $d$ with a value in $\{0, 1\}$. Variable $d$ can be used to indicate the activeness of the ReLU. The value of $d$ is set to $1$ when the input is a positive scalar, and otherwise equal to $0$.  Relaxing the domain of the variable  from discrete to a continuous closed interval $[0, 1]$ allows for continuous variability in activeness from $0$ to $1$. Below, we demonstrate that this relaxation of activation variables does not reduce the accuracy of the  solutions.

Un-rectifying ReLUs introduces an input-dependent diagonal matrix of entries $\{0, 1\}$, such that for input $\bx$,
\begin{align} \label{reLu}
\bD_{\bx} \bx = \rho \bx
\end{align}
with the following constraints:
\begin{align} \label{i}
\begin{cases}
 \mathbf{0} \leq \bD_{\bx}\bx, \\ 
(\bI-\bD_{\bx})\bx\leq\mathbf{0} \\
\bD_{\bx} = \diag(\bd) \text{ with } \bd=[d_i \in \{0, 1\}]_i.
 \end{cases}
\end{align}
In other words, $\rho \bx$ can be equivalently expressed as $\bD_{\bx} \bx$ where $\bD_{\bx}$ satisfies (\ref{i}). Input $\bx$ can run a continuum of values in the input space; however, there is a finite number of possible patterns pertaining to the diagonal elements in $\bD_{\bx}$ (no more than $2^n$, where $n$ is the dimension of $\bx$). The vector of $\{0, 1\}$-entries in $\bD_{\bx}$ 
can be regarded as a codeword of the polytope~\cite{BaraniukPowerDiagramSubdiv} at which the input is located.

If a ReLU network is un-rectified\footnote{Un-rectifying a ReLU network expresses the network as a finite number of data-dependent equations and constraints by replacing ReLU functions layer-by-layer with a finite number of data-dependent equations and constraints.}, 
then the discrete values of activation variable $\bd$ could hinder optimization, due to the fact that the discrete domain can render a combinatorial problem for which solutions cannot be efficiently derived.
In the context of optimization, variable relaxation refers to the relaxation of discrete variables by
replacing discrete values with continuous values. This kind of relaxation can be applied to (\ref{i}), as long as the solutions to the original problem are equivalent to the problem with relaxed variables. 
In the following, we demonstrate that the constraint on activation variable $d_i \in \{0, 1\}$ can be relaxed to closed interval $[0,1]$ without altering the solution for any input except the zero vector, regardless of the conversion that is applied.

The inequalities in (\ref{i}) can be converted into equations satisfying (\ref{P2a}) through the introduction of data-dependent slack variables, which would double the number of variables for a given input (compared to the size of ReLUs). For each $i$, we obtain
\begin{align} \label{islack}
\begin{cases}
d_i x_i - s_i = 0, \\
(1-d_i) x_i + t_i  = 0,\\
d_i \in \{0, 1\}, \\
t_i, s_i \geq 0.
\end{cases}
\end{align}
Slack variables $t_i$ and $s_i$ are associated with component $x_i$ of input $\bx$. 
The constraint $d_i \in \{0, 1\}$ is then relaxed to $d_i \in [0, 1]$, yielding 
\begin{align} \label{islack1}
\begin{cases}
d_i x_i - s_i = 0, \\
(1-d_i) x_i + t_i  = 0,\\
d_i \in [0, 1], \\
t_i, s_i \geq 0.
\end{cases}
\end{align}

\begin{citedlem} \label{relaxationvalid}
Let $x_i^*$, $d_i^*$, $s_i^*,  t_i^*$ be a feasible solution to (\ref{islack1}), such that $d_i^*$ is either $0$ or $1$ except when $x_i^* = 0$.
\end{citedlem}
\proof
Case $x_i^* < 0$: From the first equation in (\ref{islack1}) and $s_i^* \geq 0$, we deduce that $d_i^* \leq 0$. Meanwhile, the second equation and $t_i^* \geq 0$ lead us to deduce that $d_i^* \in [0, 1]$. To satisfy both, $d_i^*$ must equal zero. \\
Case $x_i^*  >  0$: From the first equation in (\ref{islack1}) and $s_i^* \geq 0$, we deduce that $d_i^*\geq 0$. Meanwhile, the second equation and $t_i^* \geq 0$ lead us to deduce that $d_i ^*\geq 1$. To satisfy both, $d_i^*$ must equal $1$. \\
Case $x_i^* = 0$: $d_i^* \in [0, 1]$. 

\qed

The above lemma demonstrates that the relaxation is valid.
Un-rectifying involves replacing the ReLU function with a variable in a discrete domain. 
Relaxation involves relaxing the domain of the variable, which allows it to be treated as a continuous variable. This cannot be regarded as a smoothing of the ReLU function, because most smoothing technique involve a trade-off between accuracy and computational efficiency.
Lemma \ref{relaxationvalid} indicates that this form of relaxation does not impose this kind of trade-off. From another perspective, unless un-rectification and relaxation are applied to a ReLU, the status of the function is discrete (i.e., active or in-active), with a value depending exclusively on the input of the function. Un-rectifying and relaxing a ReLU transforms the function into a constrained continuous variable. This provides an advantage in optimization, because the continuous variable can be optimized in the same way as other variables, based on its most recent input and output values.  This issue is explored in Section \ref{representationlearning}.

\section{Global convergence} \label{globalconvergence}

The following theorem sketches proof of convergence pertaining to the CGT-algorithm, which asserts that the algorithm solves (\ref{P1a})-(\ref{P3a}), provided that the primal-variable update converges to a critical point of (\ref{lagrangian1}) over $B$.

\begin{citedthm}\label{OURmaincriticalpoint}
Consider objective function $f(\bx)$ and constraints $c_i(\bx)$ on  $\bx =  (\bx^1, \bx^2) \in  B$, where $\bx^1$ is a block of unbound variables and $\bx^2$ is a block of bound variables, as defined in (\ref{P1a})-(\ref{P3a}). Suppose that (AS1)-(AS3) hold. Assume that the minimizer of the augmented Lagrangian function $L_{\mu}$ (\ref{lagrangian1})
is a critical point over $B$
of the function with fixed $\lambda^{(k-1)}$ and $\mu^{(k-1)}$ for all $k$.
Then, the CGT-algorithm can be used to solve problem (\ref{P1a})-(\ref{P3a}) and achieve global convergence wherein all of the limit points are critical points of (\ref{P1a})-(\ref{P3a}). 
\end{citedthm}
\proof

In \cite{conn1991globally} a detailed derivation is provided for the situation in which any variable lies between $l_j = 0$ and $u_j = \infty$ in order to simplify the exposition of convergent sub-sequences of iterators.  Appendix \ref{sec:globalconv} presents derivations for (\ref{P1a})-(\ref{P3a}), where some variables are unbounded and other variables are bounded within any real numbers (extended to $\infty$) $l_j$ and $u_j$ where $l_j < u_j$.  Furthermore, while we investigate additional convergent sub-sequences of iterators other than those in \cite{conn1991globally}, our proof is in line with the derivation in \cite{conn1991globally}. 
\qed

Theorem \ref{OURmaincriticalpoint} assumes that the primal variables can be updated to critical points of the augmented Lagrangian function for fixed values of dual variables. 
In the proposed DNN learning scheme, the primal-variable update is derived using the alternating minimization method of $l$ block-components:
\begin{align} \label{minfproblem}
\min_{w_i \in \mathcal D_i} L_{\mu}(w_1, \cdots, w_l).
\end{align}
This method generates the next iterator $w^{(k+1)} = (w_1^{k+1}, \cdots, w_l^{k+1})$ based on the iteration in closed convex set $\mathcal D = \mathcal D_1 \times \cdots \times \mathcal D_l$ where  $i=1, \cdots, l$:
\begin{align*}
w_{i}^{k+1} \in & \arg \min_{\eta \in \mathcal D_i} L_{\mu}(w_1^{k+1},\cdots, w_{i-1}^{k+1}, \eta, w_{i+1}^k, \cdots, w_l^k).
\end{align*}
In each iteration, $L_{\mu}$ is minimized with respect to each ``block coordinate", which is obtained in order from $w_1$ to $w_l$.

\begin{citedprop}[(Proposition 2.7.1 in \cite{Ber08})] \label{alt:stationary}
Let $L_{\mu}$ be bounded from below over closed convex set $\mathcal D= \mathcal D_1 \times \cdots \times \mathcal D_l$ and suppose that the sub-gradients at the minimum of $L_{\mu}$ are taken with respect to a block, while the other blocks are fixed. Let $\{w^{(k)}\}$ be the sequence generated using the alternating minimization method. Every limit point $\{w^{(k)}\}$ is a critical point of $L_{\mu}$ at $\mathcal D$, provided that either of the following holds: \\
(i) $L_{\mu}$ is strictly convex in block-component $i$, while the other block-components are fixed over $\mathcal D$. In other words, for each $w = (w_1, \cdots, w_l) \in \mathcal D$ and $i$, 
\begin{align*}
L_{\mu}(w_1, \cdots, w_i, \cdots, w_l)
\end{align*}
viewed as a function of $w_i$ obtains a unique minimum over closed convex set $\mathcal D_i$; or \\
(ii) Sequence $\{w_i^{(k)}\}_k$ for some $i$ is bounded, $L_{\mu}$ is a continuous function, and 
\begin{align*}
L_{\mu}(w_1, \cdots, w_i, \cdots, w_l)
\end{align*}
when viewed as a function of $w_i$ (with other block-components fixed) can be used to obtain a unique minimum over closed convex set $\mathcal D_i$.
\end{citedprop}
\qed

Theorem \ref{OURmaincriticalpoint} taken with Proposition \ref{alt:stationary} produces the following corollary.

\begin{citedcor} \label{Ourcor1}
Suppose that the assumptions pertaining to Theorem \ref{OURmaincriticalpoint} hold, and further suppose that the minimizer of the augmented Lagrangian function in the CGT-algorithm is derived using the alternating optimization method in accordance Proposition \ref{alt:stationary}. Then, all of the limit points derived using the CGT-algorithm are critical points of (\ref{P1a})-(\ref{P3a}).

\end{citedcor}

\section{Learning un-rectified networks} \label{representationlearning}

The proposed learning algorithm is based on an un-rectifying procedure, in which continuous point-wise piecewise linear activation functions are substituted layer-by-layer into a non-linear optimization problem with a finite number of data-dependent equations and constraints.
All in-equality constraints are made equalities through the introduction of data-dependent slack variables and all activation variables of ReLUs are relaxed to the closed interval $[0, 1]$. The resulting un-rectifying representation of the following $L$-layer ReLU network is given in 
Section \ref{learningL}:
\begin{align*}
\by = \mathcal M_L(\bx) =M_L (\rho_{L-1} M_{L-1}(\cdots (\rho_2M_2(\rho_1M_1(\bx)))\cdots)), 
\end{align*}
where $M_n(\bx)=\bW_n \bx +\bb_n$, $\bW_n \in \R^{N_n \times N_{n-1}}$ are weight matrices, $\bb_n \in \R^{N_n}$ are biases, $\bx \in \R^{N_0}$ is the input, and $\rho (t)  := \max(0,t)$ for $t\in\R$.
Throughout the rest of this paper, $M_n\bx$ is used as shorthand for $M_n(\bx)$.

\subsection{Learning two-layer networks}

To elucidate the un-rectifying procedure, we consider the regression problem used to learn a two-layer representation $\mathcal M_2$ from $N$ training data $\{(\bx_j \in \R^{N_0},\by_j \in \R^{N_2})\}_{j=1}^N$.
By imposing the Frobenius norm on the weight matrices to stabilize the outputs with respect to input perturbation\cite{Wen19}, the least squares regression problem becomes 
\begin{align} \label{2objective}
\sum_{j}\|\by_j - \mathcal M_2 \bx_j\|^2 + \frac{c_1}{2}(\|\bW_1\|_F^2 + \|\bW_2\|_F^2),
\end{align}
where $c_1 > 0$ is a parameter. 
We un-rectify the ReLU in  $\mathcal M_2$ by replacing $\rho$ with data-dependent diagonal matrices $\bD_j$ and add the term $\frac{c_2}{2} \sum_j \|\bD_j\|_F^2$ to (\ref{2objective}) in which the value of $c_2$ is small enough that its effect on the solution of (\ref{2objective}) is negligible. Including the term produces a unique closed-form solution of variables in $\bD_j$, where $\bD_j$ is defined as \begin{align} \label{ReLUequ}
\bD_j M_1\bx_j = \rho M_1\bx_j
\end{align}
and the $i$-th diagonal entry of $\bD_j$ is denoted as 
\begin{align}\label{alt:ReLUdefineD}
d_{ji}=
\begin{cases}
1 	&\text{ if }( M_1 \bx)_i = (\bW_1\bx_j + \bb_1)_i > 0,\\
0    &\text{ else}.
\end{cases}
\end{align}
We introduce data-dependent slack variables to convert inequality constraints derived from un-rectifying ReLUs into equality constraints with the aim of obtaining a constrained optimization problem seeking $\bW_1$, $\bb_1$, $\bW_2$, $\bb_2$, $\{d_{ji}\}$, $\{s_{ji}\}$, $\{t_{ji}\}$ to locally minimize the following:
\begin{align*} 
& 
\begin{cases}
\sum_j\|\by_j - M_2\bD_jM_1\bx_j\|^2+ \frac{c_1}{2}(\|\bW_1\|_F^2 + \|\bW_2\|_F^2) \\ 
\hspace{1.2in} + \frac{c_2}{2} \sum_j \|\bD_j\|_F^2,  \\
\text{ subjected to}
\begin{cases}  
d_{ji}  (M_1 \bx_{j})_i  - s_{ji} = 0, \\
(1-d_{ji}) (M_1 \bx_{j})_i + t_{ji}  = 0,\\
d_{ji} \in [0, 1], \\
t_{ji}, s_{ji} \geq 0.
\end{cases}
\end{cases}
\end{align*}
In accordance with the three-splitting approach, the introduction of data-dependent vectors $\bu_j = [u_{ji}]_i = M_1 \bx_j$ and $\bv_j =[v_{ji}]_i = \bD_j \bu_j$ allow re-expression of the above as \begin{align*} 
\text{(P2)} & 
\begin{cases}
\text{Seek for $\bW_1, \bb_1, \bW_2, \bb_2, \{d_{ji}\}, \{u_{ji}\}, \{v_{ji}\}, \{s_{ji}\}, \{t_{ji}\}$ that is a local minimizer to} \\
\sum_j\|\by_j - M_2 \bv_j\|^2+ \frac{c_1}{2}(\|\bW_1\|_F^2 + \|\bW_2\|_F^2) + \frac{c_2}{2} \sum_j \|\bD_j\|_F^2,  \\
\text{ subjected to}
\begin{cases}
v_{ji} = d_{ji} u_{ji}, \\
u_{ji} = (M_1 \bx_j)_i, \\  
d_{ji} u_{ji} - s_{ji} = 0, \\
(1-d_{ji}) u_{ji} + t_{ji} = 0, \\
d_{ji} \in [0, 1], \\
t_{j,i}, s_{j,i} \geq 0.
\end{cases}
\end{cases}
\end{align*}
With an exclusive focus on the equality constraints in (P2), we formulate the augmented Lagrangian function of (P2) to be analyzed using the CGT-algorithm. The primal variables include unbounded variables ($\bW_2$, $\bb_2$, $\bW_1$, $\bb_1$, $\{u_{ji}\}$,  $\{v_{ji}\}$) and simple bounded variables ($\{t_{j,i} \geq 0\}, \{s_{j,i} \geq 0\}, \{d_{j,i} \in [0, 1]\}$). As demonstrated in the Appendix \ref{appendixprimalvariable}, the primal variables (unbounded and simple bounded variables) can be updated efficiently in (reverse) from layer two to layer one. The primal variables are alternatively updated to reach a local minimum of the 
augmented Lagrangian function when dual variables are fixed. This conclusion is a consequence of Proposition \ref{alt:stationary}, since a unique minimizer can be derived for the updating of any primal variable. This is evidenced by the fact that the augmented Lagrangian function is a strongly quadratic convex function when viewed as a function of any of the above variables with the other variables fixed.

The objective function and equality constraints in (P2) are twice-continuously-differentiable functions. Thus, (AS1) holds. (AS2) is a pragmatic assumption based on the fact that the iterators must be bounded in practice when an algorithm is executed. The size of Jacobian matrix of (P2) is $J_1 \times J_2$ with $J_2 > J_1$ everywhere, where $J_1= 4 N N_1$ is the number of equality constraints and $J_2$ is the number of primal variables, which is $5 NN_1 + N_0 N_1 + N_1 N_2 + N_1 + N_2$ \footnote{$j$ and $i$ in (P2) runs from $1$ to $N$ (the number of training data) and $1$ to $N_1$ (the number of rows in $\bW_1$). The sizes of $\bW_1$, $\bW_2$, $\bb_1$, and $\bb_2$ are $N_1 \times N_0$, $N_2 \times N_1$, $N_1$ and $N_2$, respectively}.  Directly from Corollary \ref{Ourcor1}, we obtain the following conclusion:

\begin{citedcor} \label{2-layerconclusion}
Suppose that (AS3) holds with any limit $\bx^*$\footnote{This means that the rank of the Jacobian matrix $ \bA(\bx^*)|_J$ is $4N N_1$, where $J$ refers to the indices of unbound variables and bound variables, whose limits are not in the bounds, $N$ is the number of training data, and $N_1$ is the number of ReLUs in $\mathcal M_2$.} when (P2) is solved using the CGT-algorithm. 
Then, the two-layer ReLU-regression problem (P2) can be solved using the augmented Lagrangian approach with global convergence using the CGT-algorithm, in which the updating of primal variables is conducted using the alternating minimization method where the variables are updated in reverse order from layer two to layer one.
\end{citedcor}

\subsection{Learning L-layer networks} \label{learningL}

We streamlined the derivation by resolving the general L-layer regression problem as in the two-layer case.
We consider the regression problem for learning an L-layer ReLU network $\mathcal M_L\colon \R^{N_0} \to \R^{N_L}$ from $\{(\bx_j, \by_j) | \bx_j \in \R^{N_0}, \by_j \in \R^{N_L}\}_{j=1}^N$ input data in which
{\small{
\begin{align*}
\mathcal M_L \bx = 
\bW_L\rho(\bW_{L-1}\cdots (\bW_2 \rho (\bW_1 \bx + \bb_1) + \bb_2) + \cdots )+ \bb_{L}
\end{align*}
where $\rho$s are the ReLU activation functions and $M_i \bx = \bW_i \bx+ \bb_i$ with $\bW_i \in \R^{N_i \times N_{i-1}}$.
After $\rho$s are un-rectified by the introduction of data-dependent diagonal matrices 
 $\{\bD^1_j, \cdots, \bD^L_j \}$ and constraints for input $\bx_j$, regularization of the Frobenius norm of weighting matrices, and introduction of quadratic term $\sum_{j,k}\|\bD^k_j\|^2$ allows us to formulate the problem as follows:
 
\begin{align*} 
 & 
\begin{cases}
\sum_{j=1}^N \|\by_j - \mathcal  M_L \bx_j\|^2 
 + \frac{c_1}{2}\sum_{l=1}^L\|\bW_l\|_F^2 \\ \hspace{1.2in}  +\frac{c_2}{2} \sum_{j=1}^N \sum_{k=1}^{L-1}\|\bD^k_j\|_F^2 \\
\text{ subjected to}
\begin{cases}
-d_{ji}^k ( M_k\bD^{k-1}_j\cdots \bD^1_j  M_1 \bx_j)_i + s_{ji}^k= 0, \\
(1-d_{ji}^k)(  M_k\bD^{k-1}_j\cdots \bD^1_j M_1 \bx_j)_i +t_{ji}^k= 0 ,\\
d_{ji}^k \in [0, 1], \\
s_{ji}^k, t_{ji}^k \geq 0,
\end{cases}
\end{cases}
\end{align*}
where $j$ is data index ($j=1, \cdots, N$), $k$ is layer index ($k=1, \cdots, L-1$), $i$ is layer-dependent ($i = 1, \cdots, N_k$), $c_2$ is a very small value (introduced so that the unique solution to $\{d_{ji}\}$ can be efficiently derived), and $\bD^k_j =\diag([d^k_{ji}]_i)$. 
Following the three-splitting method, data-dependent vectors $\bu_j^k = [u^k_{ji}]_i = M_k \bD^{k-1}_j\cdots \bD^1_j  M_1 \bx_j$ and $\bv_j^k =[v^k_{ji}]_i = \bD_j^k \bu_j^k$ are introduced. The above problem can now be expressed as the search for
$\{\bW_k\}_k$, $\{\bb_k\}_k$, $\{\bD^k_j\}_{k,j}$, $\{s^k_{ji}\}_{i,j,k}$ and $\{t^k_{ji}\}_{i,j,k}$ to locally minimize the following:
 \begin{align*} 
\text{(PL)} & 
\begin{cases}
\sum_{j=1}^N \|\by_j - M_L \bv_j^{L-1}\|^2 
 + \frac{c_1}{2}\sum_{l=1}^L\|\bW_l\|_F^2 + \frac{c_2}{2} \sum_{j=1}^N \sum_{k=1}^{L-1}\|\bD^k_j\|_F^2 \\
\text{ subjected to}
\begin{cases}
v_{ji}^{k} = d_{ji}^{k} u_{ji}^{k}, \\
u_{ji}^{k} = (M_k \bv_j^{k-1})_i \text{ and } \bv_{j}^{0} = \bx_{j}, \\  
d_{ji}^{k} u_{ji}^{k} - s_{ji}^{k} = 0, \\
(1-d_{ji}^{k}) u_{ji}^{k} + t_{ji}^{k} = 0, \\
d_{ji}^{k} \in [0, 1], \\
t^k_{ji}, s^k_{ji} \geq 0,
\end{cases}
\end{cases}
\end{align*}

Analogous to the two-layer case, the augmented Lagrangian function that can be analyzed using the CGT-algorithm is obtained from the objective function and the equality constraints.
The primal variables include unbounded variables ($\{\bW_k, \bb_k\}$, $\{u^k_{ji}, v^k_{ji}\}$) and simple bounded variables ($\{t^k_{j,i} \geq 0\}, \{s^k_{j,i} \geq 0\}, \{d^k_{j,i} \in [0, 1]\}$). As demonstrated in Appendix \ref{appD}, the primal variables can be efficiently updated in reverse order from later layers to earlier layers. The primal variables are alternatively updated to reach a local minimum of the 
augmented Lagrangian function in which the dual variables are fixed. This conclusion is a consequence of Proposition \ref{alt:stationary}, since unique minimizer can be derived for the updating of any primal variable. This is evidenced by the fact that the augmented Lagrangian function is a strongly quadratic convex function when viewed as a function of any of the above variables with the other variables fixed.

The objective function and equality constraints in (PL) are twice-continuously-differentiable functions. Thus, (AS1) holds. (AS2) is a pragmatic assumption based on the fact that the iterators must be bounded in practice when an optimization algorithm is executed. Analogous to the two-layer case, global convergence in  learning L-layer ReLU networks using the CGT-algorithm can be achieved as long as the rank of the Jacobian matrix at $\bx^*$ satisfies (AS3), where $\bx^*$ is a limit point when (PL) is solved using the CGT-algorithm. The size of Jacobian matrix for (PL) is $J_1 \times J_2$ with $J_2 > J_1$ everywhere, where $J_1= 4 (L-1) N \sum_{l=1}^{L-1} N_l$ is the number of equality constraints and $J_2$ is the number of primal variables, derived as follows: $ 5 (L-1) N \sum_{l=1}^{L-1} N_l + \sum_{i=1}^L N_{i-1} N_i  +\sum_{i=1}^L  N_l$.  The following conclusion follows from Corollary \ref{Ourcor1}:

\begin{citedcor} \label{l-layerconclusion}
Suppose that (AS3) holds with limit $\bx^*$ and the rank of the Jacobian matrix $ \bA(\bx^*)|_J$ is $4 (L-1) N \sum_{l=1}^{L-1} N_l$, where $J$ refers to the indices of unbound variables and bound variables whose limits are not in the bounds, $N$ is the number of training data, and $\sum_{l=1}^{L-1} N_l$ is the number of ReLUs in $\mathcal M_L$. The L-layer ReLU-regression problem (PL) can then be solved using the augmented Lagrangian approach with global convergence using the CGT-algorithm, in which updating of the primal variable is conducted using the alternating minimization method with variables updated in reverse order from layer $L$ to layer $1$.
\end{citedcor}

The CGT-algorithm for (PL) is presented in Algorithm \ref{alder-al}. The algorithm learns an optimal L-layer representation from the data by solving the constrained optimization problem derived using the un-rectifying method.

\begin{algorithm}[!h]
\caption{CGT-algorithm solving (PL) }
\label{alder-al}
\algsetup{indent=2em}
\begin{algorithmic}[1]
\STATE  INPUT: the number of layers, $L$; the training data $\{(\bx_j \in \R^{N_0}, \by_j\in \R^{N_L})\}_{j=1}^N$ ; parameters $c_1$ and $c_2$; penalty parameter $\rho_1^{(0)} = \rho_2^{(0)}=\rho_3^{(0)}=\rho_4^{(0)}= \rho^{(0)}  > 0$; $\bar \mu^{(0)} = \{\{(\mu_{1ji}^l)^{(0)}= 0\} \cup  \{(\mu_{2ji}^l)^{(0)}= 0\}\} \cup \{(\mu_{3ji}^l)^{(0)}= 0\} \cup \{(\mu_{4ji}^l)^{(0)}= 0\}\}$ (the Lagrangian multipliers of equality constraints in (PL)) where $j = 1, \cdots, N$; $l=1,\cdots, L-1$; and $i$ is layer-dependent with $i=1, \cdots, N_{l}$; and stopping conditions $\omega^*$ and $\eta^*$, and $Iter$ (the maximum number of iterations).
\STATE Let $\alpha$ denote primal variables and let $\bc$ denote the vector of all equality constraints in (PL).
\STATE Set $k = 0$, $\omega_0 = \omega^{(0)} = 1$, $\eta_0=\eta^{(0)}= 1$, and $\tau = 0.01$, as suggested in \cite{conn1991globally}.
\REPEAT
\REPEAT 
\STATE $\alpha^{(k)} \leftarrow$ Update primal variables $\alpha$ using Algorithm \ref{funrec} with fixed Lagrangian multipliers $\bar \mu^{(k)}$.
\UNTIL $\|\bP(\alpha^{(k)}, \nabla_{\alpha} L_{\rho^{-1}}^{(k)}) \| \leq \omega^{(k)}$
\IF {$\|\bc(\alpha^{(k)}) \|\leq  \eta^{(k)}$ }
\STATE [Test for convergence and update Lagrangian multiplier estimates].
\IF { $\|\bP(\alpha^{(k)}, \nabla_{\alpha} L_{\rho^{-1}}^{(k)}) \| \leq \omega^*$ \text{and} $\|\bc(\alpha^{(k)}) \| \leq \eta^*$ }
\STATE Return $\bW_l^{(k)}$, $\bb_l^{(k)}$ with $l=1, \cdots, L$.
\ENDIF
\STATE [Update dual variables by using (\ref{update_dual})].
\STATE $\rho^{(k+1)} = \rho^{(k)}$; $\beta = \min((\rho^{(k+1)})^{-1}, 0.1)$;  $\omega^{(k+1)} = \omega^{(k)} \beta$; $\eta^{(k+1)} = \eta^{(k)} \beta^{0.9}$.
\ELSE 
\STATE [Reduce the penalty parameter $\rho^{-1}$  (or increase $\rho$) ]. 
\STATE $(\rho^{(k+1)})^{-1} = \tau  (\rho^{(k)})^{-1}$.
\STATE $\bar \mu^{(k+1)} =\bar\mu^{(k)}$; $\beta = \min((\rho^{(k+1)})^{-1}, 0.1)$.
\STATE $\omega^{(k+1)} = \omega_0 \beta$ and $\eta^{(k+1)} = \eta_0 \beta^{0.1}$.
\ENDIF
\STATE Increase $k$ by one.
\UNTIL $k \ge Iter$
\end{algorithmic}
\vspace{1pt} 
{\hrule height0.5pt} 
\vspace{6.5pt}
\noindent \textbf{Remark 1.} The value of $\rho^{(k)}$ is a non-decreasing function of $k$.
The updating of dual variables of (\ref{2aug00}) (the augmented Lagrangian function of (PL)) is performed as follows:
\begin{align} \label{update_dual}
\begin{cases}
(\mu_{1ji}^l)^{(k+1)} \leftarrow (\mu_{1ji}^l)^{(k)} + \rho_1^{(k)} ((v_{ji}^l)^{(k)} - (d_{ji}^{l})^{(k)}(u_{ji}^{l})^{(k)}),\\
(\mu_{2ji}^l)^{(k+1)} \leftarrow (\mu_{2ji}^l)^{(k)} + \rho_2^{(k)} ((u_{ji}^l)^{(k)} - (\bW_{l}^{(k)}(\bv_{j}^{l-1})^{(k)}+\bb_{l}^{(k)})_i),\\
(\mu_{3ji}^l )^{(k+1)}\leftarrow (\mu_{3ji}^l)^{(k)}+  \rho_3^{(k)} ((d_{ji}^l)^{(k)}(u_{ji}^{l})^{(k)} - (s_{ji}^{l})^{(k)}),\\
(\mu_{4ji}^l )^{(k+1)}\leftarrow (\mu_{4ji}^l)^{(k)}+  \rho_4^{(k)} ((1-(d_{ji}^l)^{(k)})(u_{ji}^{l})^{(k)} + (t_{ji}^{l})^{(k)}). \\
\end{cases}
\end{align}
\noindent\textbf{Remark 2.}The initial values are as follows: $\rho_1$ and $\rho_2$ ($1$),  $\rho_3$ and  $\rho_4$ ($100$), $c_1$ ($0.001$), and $c_2$ ($10^{-6}$).
\end{algorithm}

\section{Compressed sensing recovery and experiments} \label{exp}

Compressed sensing (CS) theory dictates that a signal exhibiting sparsity in some transform domain can with high probability be reconstructed from far fewer measurements. CS has shown considerable promise in a variety of applications, including but not limited to low-cost on-sensor image acquisition, wireless tele-monitoring, and accelerating magnetic resonance imaging.

For a given $m\times n$ sensing matrix $\bA$ in which $m < n$ and measurement $\by \in \R^m$, the compressed sensing recovery problem involves recovering the sparsest signal $\bx \in \R^n$ 
by solving the following optimization problem:
\begin{align} \label{csproblem}
\begin{cases}
\min_{\bx} \|\bx \|_0 \\
\| \by - \bA \bx \|_2^2 \leq \epsilon
\end{cases}
\end{align}
where $\epsilon > 0$ is a given parameter. This is an ill-posed inverse problem, as $n > m$, where the CS ratio is defined as $\frac{m}{n}$. 
Theoretically it is possible to develop algorithms capable of deriving a solution to (\ref{csproblem}); however, the actual solution process comprises numerous updating iterations.
Each update solves an optimization sub-problem typically involving matrix-matrix multiplication and matrix inversion operations. This can hamper applications using devices with limited computing resources and the optimizations of large-scale systems. Deep neural networks (DNNs) provide an alternative perspective by which to overcome these difficulties. Instead of solving the problem using an inverse method, DNN-based CS recovery algorithms forwardly infer sparse vectors by learning the inverse mapping from $\R^m$ to $\R^n$. 

In the current study, we adopted the approach proposed in \cite{mousavi2017learning} in which a DNN learns a deep (fully-connected) ReLU network mapping sparse vectors from measurements. 
Our training data is $(\bA^+\bY, \bX)$ where $\bX$ is the array with sparse vectors in columns, $\bA^+$ is the pseudo-inverse of $\bA$, $\bA \bA^+ =\bI$, and arrays $\bY$ and $\bX$ are related to each other via $\bY = \bA \bX$.
ReLU network $\mathcal M_L$ is learned with the aim of optimizing the least-squares regression problem as follows:
\begin{align}
\| \bX - \mathcal M_L \bA^+ \bY \|_F^2.
\end{align}
$\mathcal M_L$ is un-rectified, formulated as (PL), and then learned using Algorithm \ref{alder-al}. In testing, the vector of $\by$ with respect to $\bA$ is obtained via forward-inference using
\begin{align} \label{forwardinference}
\hat \bx = \mathcal M_L \bA^+ \by.
\end{align}
We conducted a series of experiments applying the proposed scheme to the MNIST dataset and natural images. The performance of the proposed scheme was compared with that of the Adam optimizer (Matlab builtin function $trainNetwork$), which is a state-of-the-art implementation of the back-propagation algorithm. We also assessed the block-coordinate-descent method (BCD), which optimizes the three-splitting formulation~\cite{zeng2018global} of DNN learning~\footnote{The code of BCD is publicly available at ``https://github.com/timlautk/BCD-for-DNNs-PyTorch"}.
All of the algorithms learnt an eight-layer ReLU network, $\mathcal M_8$, using the same sets of training data and the same set of testing data.

\subsection{Initialization phase}

Learning a ReLU network is a non-convex optimization problem. This means that the (local) optimal solution produced by a learning algorithm is subject to the initial weights of the network. The initial weights are typically obtained via sampling from an independently identically distributed (i.i.d) zero-mean Gaussian random variable of a given variance. In order to have a fair basis for comparison, we began by conducting experiments aimed at determining the variance of an i.i.d. zero-mean Gaussian random variable for initial weights that work well with all the selected methods. This was achieved using the following simulation:

Let $Y \in \R^{784 \times 60,000}$ denote the output of a two-layer ReLU network using the MNIST dataset as an input. Let $X \in \R^{784\times60,000}$ denote the input and let $\sigma$ denote the ReLU activation functions. Note that $Y=\bar W_2 \sigma \bar W_1 X$,
where $\bar W_1 \in \R^{784\times784}$ and $\bar W_2\in \R^{784\times784}$ are Gaussian matrices.  Each element in a matrix was sampled from an i.i.d. Gaussian random variable with the mean set at zero and standard deviation set at one. A six-layer ReLU network $\mathcal M_6$ with weight matrices $\bW_i \in \R^{784 \times 784}$ is then learned from the objective $\|X-\mathcal M_6 Y\|_F^2$, where the training input is $Y$ and output is $X$. Note that regularization terms specific to each learning method were added to the objective to optimize respective learning performance. We then compared the performance of the networks that emerged when the initial weights were sampled from the i.i.d. zero-mean Gaussian random variable with the standard deviation of the weights set at $0.1$, $0.05$ \cite{glorot10, he15}, and $0.01$, respectively. The BCD was run using the default settings and the Adam optimizer was run with the  learning rate set at $0.001$. The networks were learned using batch data.

Figure \ref{fig:mnist_simu} compares the perceptual quality of reconstruction of an image from the MNIST dataset. As shown in the figure, only the images with the standard deviation set at $0.01$ are recognizable, regardless of the learning method. The proposed scheme was less susceptible to the initial weights than were the other methods. Figure \ref{fig:mnist_err} compares the performances of the methods in terms of average mean squared error (MSE) in the reconstructed images versus the number of epochs (iterations).  As shown in the figure, only the proposed un-rectifying method was robust to the initial standard derivation of the weights, wherein the performance curves decreased smoothly to less than $0.02$ average MSE per pixel when the number of iterations exceeded $100$.  However, when the initial standard derivation was set at $0.01$, the average MSE decreased gradually with an increase in the number of iterations (bottom row of Figure \ref{fig:mnist_err}).
Setting standard deviation to $0.01$ yielded satisfactory results, regardless of which method was employed. 
%

The following experiment in image classification involved learning a network capable of predicting a number from an input image. We employed the one-hot encoding method to encode images of digits (from $0$ to $9$) in the MNIST dataset. For instance, an image of the digit $1$ is the input and the one-hot encoding output is $[0, 1, 0, 0, 0, 0, 0, 0, 0, 0]^\top$. Likewise, an image of the digit $9$ is represented as the one-hot encoding output of $[0, 0, 0, 0, 0, 0, 0, 0, 0, 1]^\top$. We let $X$ denote the input vectors of the MNIST dataset and $Y$ denote the output one-hot vectors. A six-layer ReLU network $\mathcal M_6$ learns $(X, Y)$ by minimizing $\|Y-\mathcal M_6 X\|_F^2$ with additional regularization terms, depending on which learning method is adopted. The dimensions of $W_1$ to $W_5$ were set at $784\times784$ and $W_6$ was set at $10\times784$. The initial weights of $W_i$ were Gaussian matrices of i.i.d. Gaussian random variables with zero means and initial standard deviations set at $0.1$, $0.05$, and $0.01$, respectively. The initial biases of affine mappings were set at $0$. 

If the locations of the maximum values between the output of a network and the desired one-hot encoding of an input are matched, then the input is correctly classified and the accuracy is set at $1$; otherwise, it is misclassified and the accuracy is set at zero. The accuracy of a network is determined by averaging the  accuracy of the network with respect to input vectors. Figure \ref{fig:mnist_class} presents the average accuracy of networks derived using various numbers of epochs (iterations) and initial weights. From Figures \ref{fig:mnist_simu},\ref{fig:mnist_err}, and \ref{fig:mnist_class}, we conclude that the robustness of the proposed un-rectifying method to the initial weight setting is superior to that of other methods. Setting standard deviation to $0.01$ yielded satisfactory results, regardless of which method was employed.

\begin{figure}[!ht]
\begin{center}
\mbox{
\fbox{\includegraphics[width=0.2\textwidth]{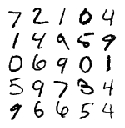}}
} \\
\vspace{0.2in}
\mbox{
\fbox{\includegraphics[width=0.2\textwidth]{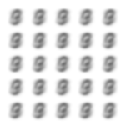}}
\fbox{\includegraphics[width=0.2\textwidth]{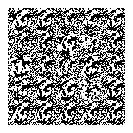}}
\fbox{\includegraphics[width=0.2\textwidth]{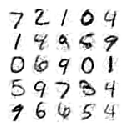}}
}
\mbox{
\fbox{\includegraphics[width=0.2\textwidth]{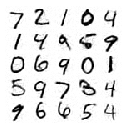}}
\fbox{\includegraphics[width=0.2\textwidth]{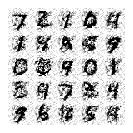}}
\fbox{\includegraphics[width=0.2\textwidth]{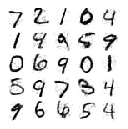}}
}
\mbox{
\subfigure[Adam]{\fbox{\includegraphics[width=0.2\textwidth]{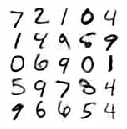}}}
\subfigure[BCD]{\fbox{\includegraphics[width=0.2\textwidth]{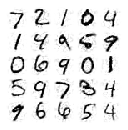}}}
\subfigure[Proposed]{\fbox{\includegraphics[width=0.2\textwidth]{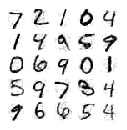}}}
}
\caption{Comparison of image recovery performance based on reconstruction results obtained using (a) Adam, (b) BCD, and (c) proposed methods with learning rate of Adam optimizer set at $0.001$, parameters of BCD set at defaults, and the initial settings of the proposed method set according to Remark 2 in Algorithm \ref{alder-al}. (first row): the original image; standard deviations of (second row) $0.1$, (third row) $0.05$, and (fourth row) $0.01$.
}
\label{fig:mnist_simu}
\end{center}
\end{figure}

\begin{figure}[!ht]
\begin{center}
\mbox{
{\includegraphics[width=0.32\textwidth]{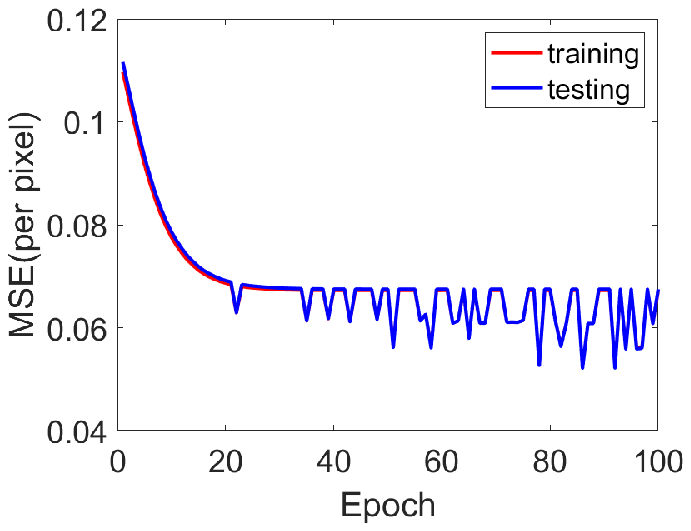}}
{\includegraphics[width=0.32\textwidth]{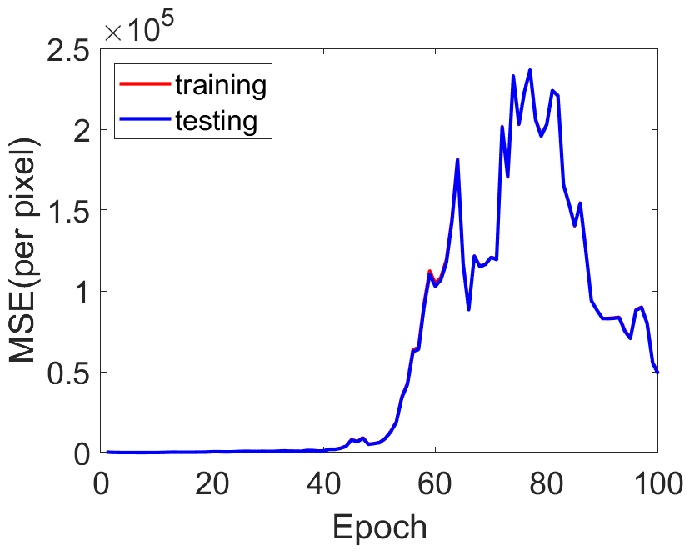}}
{\includegraphics[width=0.32\textwidth]{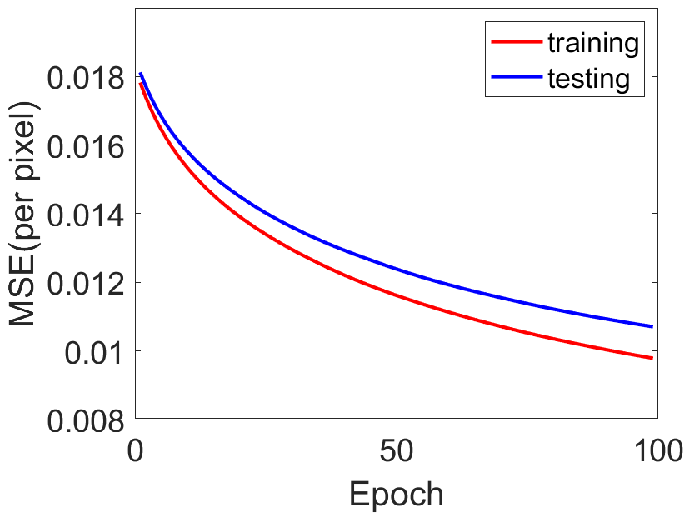}}
}
\mbox{
{\includegraphics[width=0.32\textwidth]{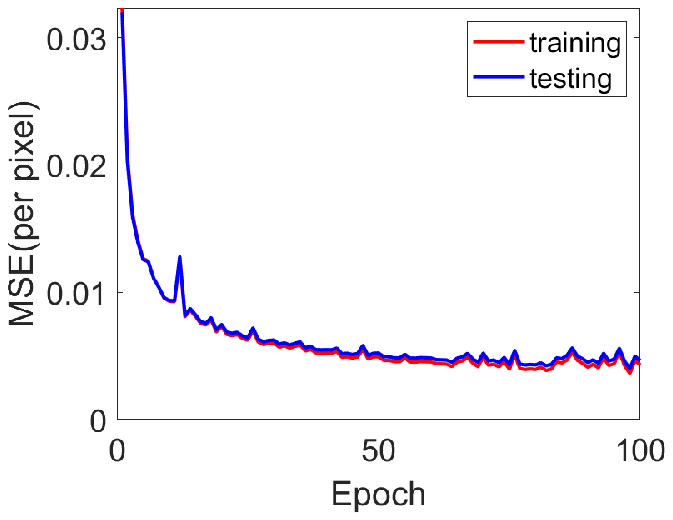}}
{\includegraphics[width=0.32\textwidth]{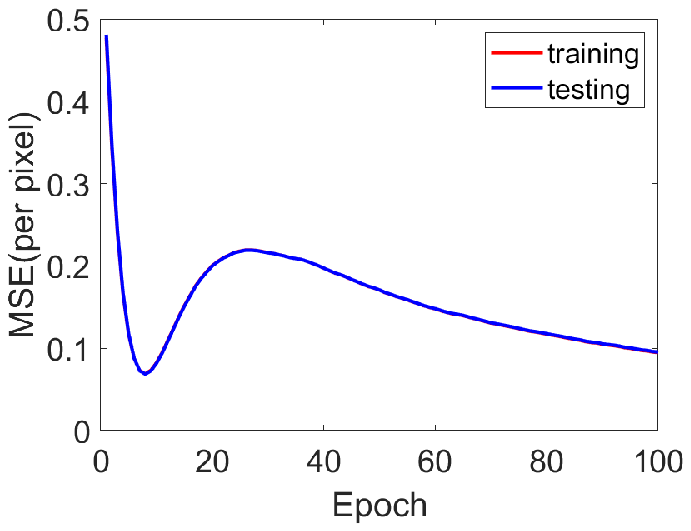}}
{\includegraphics[width=0.32\textwidth]{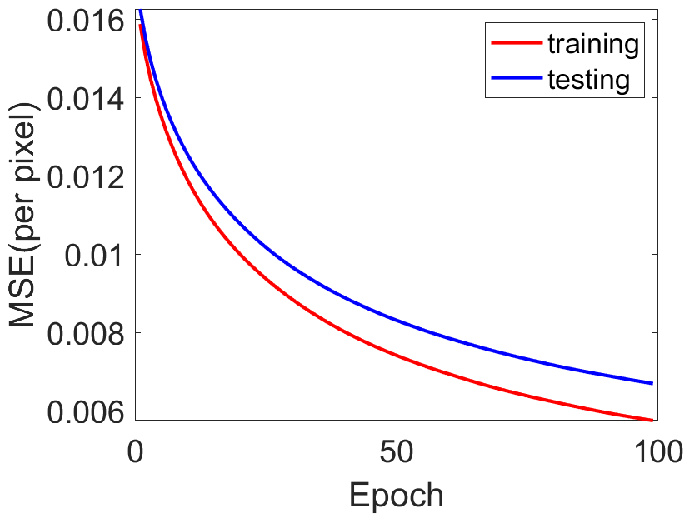}}
}
\mbox{
\subfigure[Adam]{\includegraphics[width=0.32\textwidth]{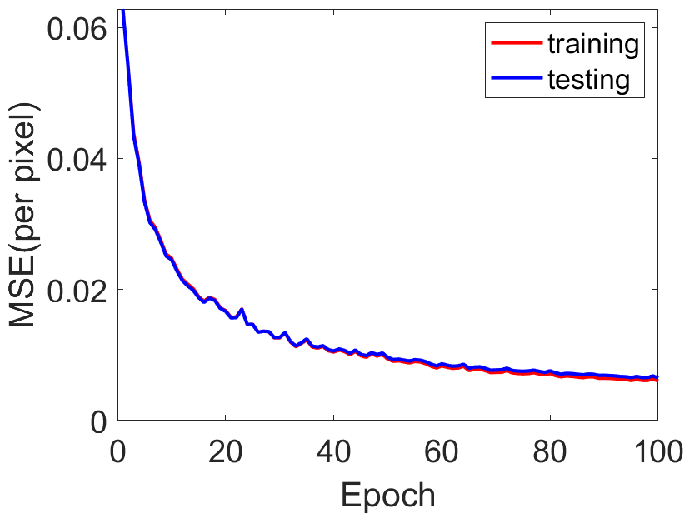}}
\subfigure[BCD]{\includegraphics[width=0.32\textwidth]{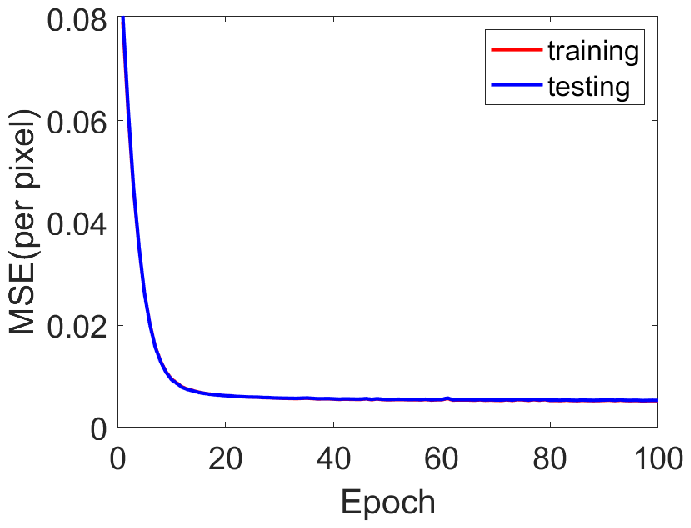}}
\subfigure[Proposed]{\includegraphics[width=0.32\textwidth]{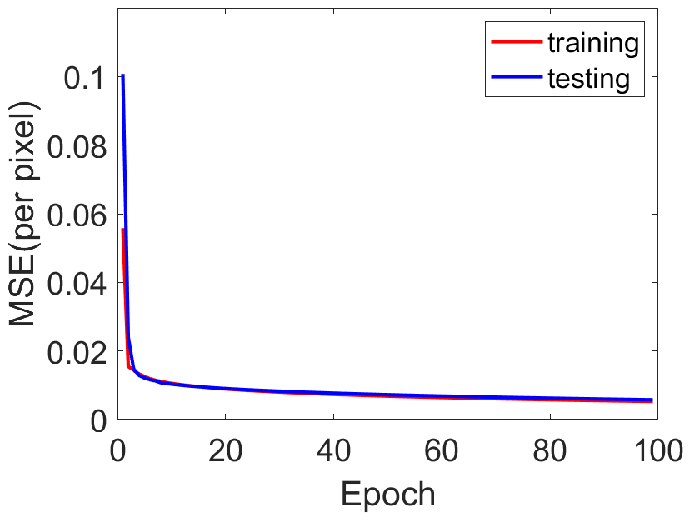}}
}
\caption{Comparison of image recovery performance of (a) Adam, (b) BCD, and (c) proposed methods in terms of average MSE with learning rate of Adam optimizer set at $0.001$, parameters of BCD set at defaults, and initial settings of the proposed  method set according to Remark 2 in Algorithm \ref{alder-al}. Standard deviation of initial weights set at (top row) $0.1$, (middle row) $0.05$, and (bottom row) $0.01$. 
The Adam method presents fluctuations after $40$ iterations with standard deviation set at $0.1$. The
BCD method does not decrease smoothly as the number of iterations increases with standard deviations $0.1$ and $0.05$. There is a large error in the top curve in (b) after $40$ iterations. With the standard deviation set at $0.05$, the results of all methods were acceptable. }
\label{fig:mnist_err}
\end{center}
\end{figure}

\begin{figure}[!ht]
\begin{center}
\mbox{
{\includegraphics[width=0.32\textwidth]{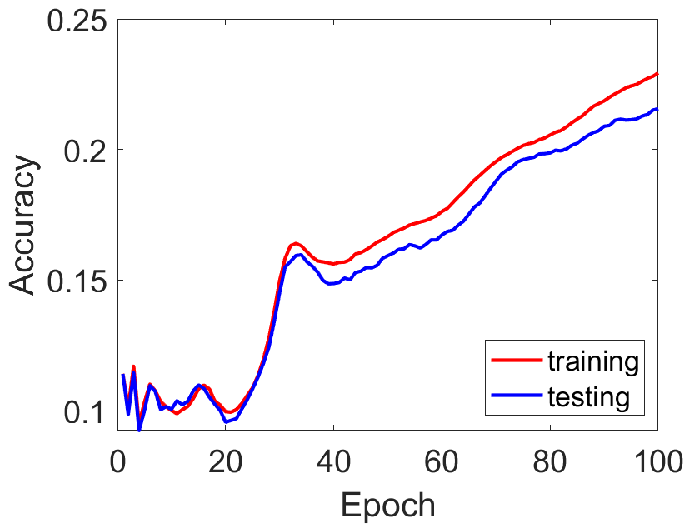}}
{\includegraphics[width=0.32\textwidth]{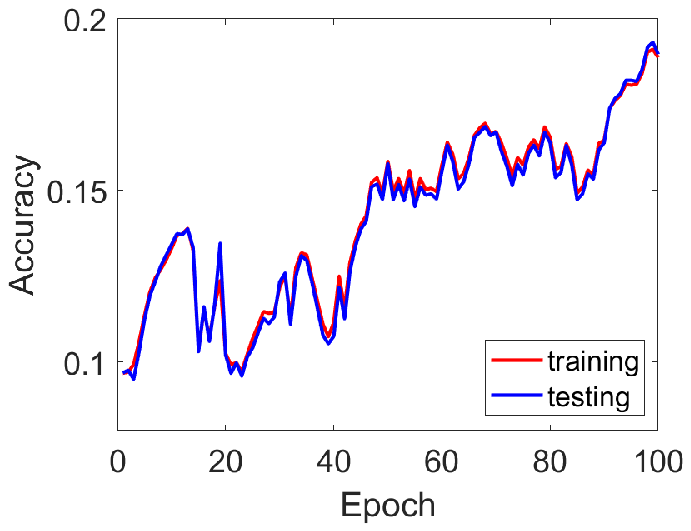}}
{\includegraphics[width=0.32\textwidth]{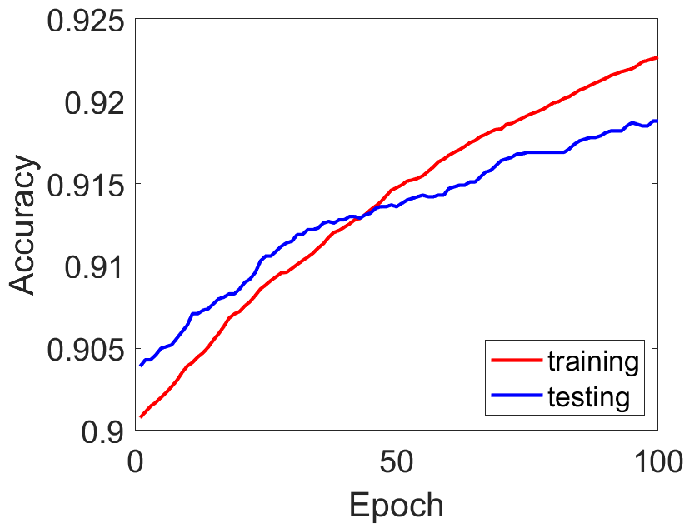}}
}
\mbox{
{\includegraphics[width=0.32\textwidth]{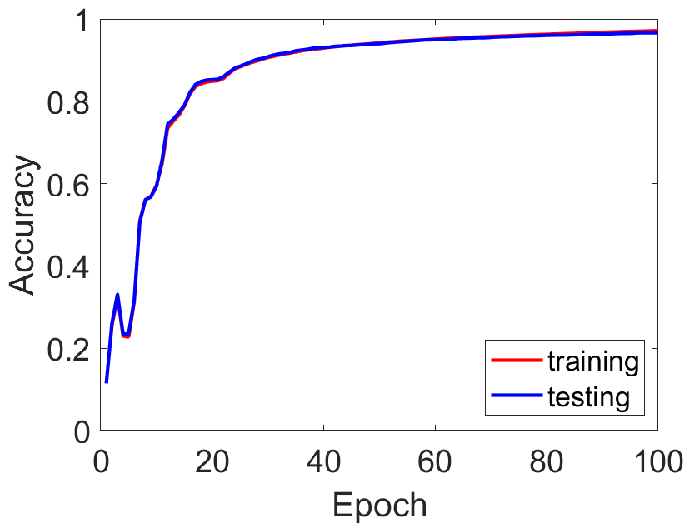}}
{\includegraphics[width=0.32\textwidth]{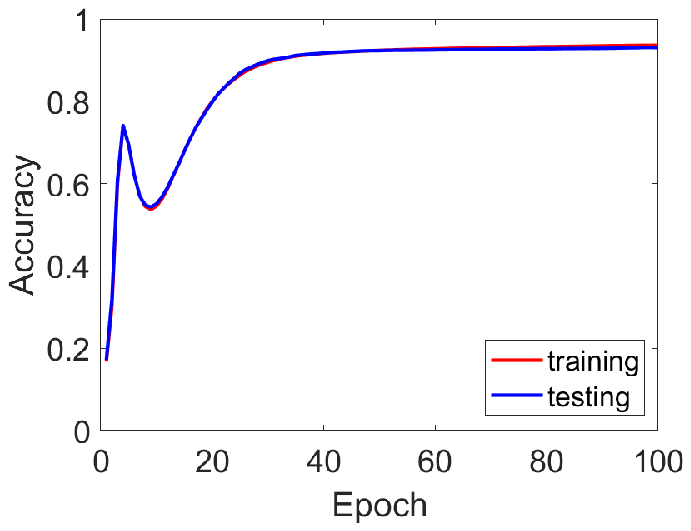}}
{\includegraphics[width=0.32\textwidth]{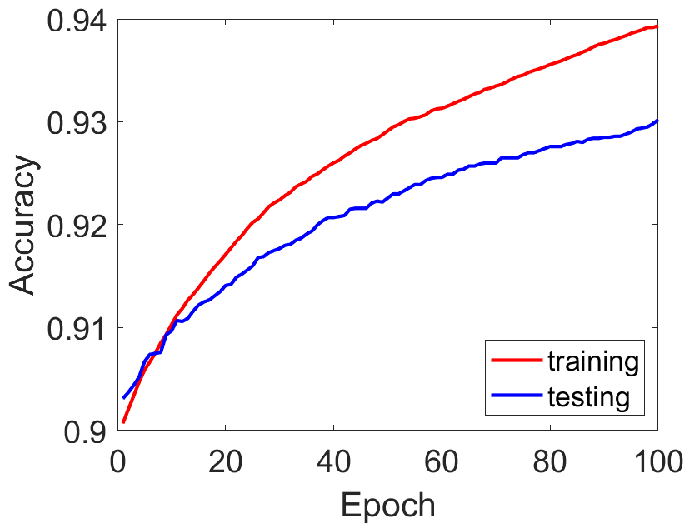}}
}
\mbox{
\subfigure[Adam]{\includegraphics[width=0.32\textwidth]{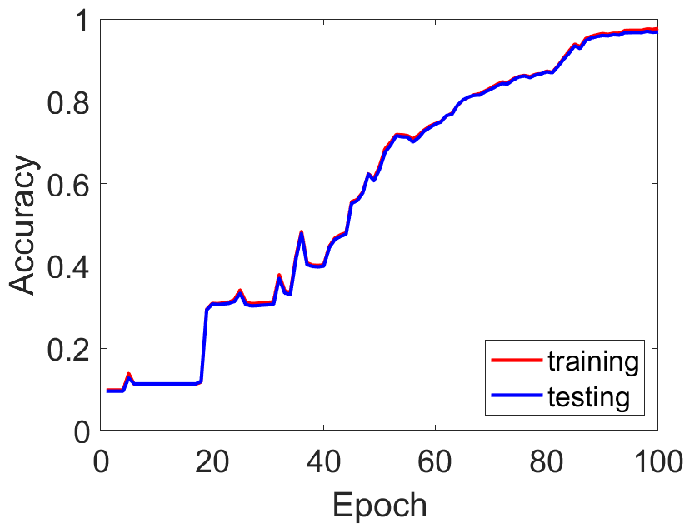}}
\subfigure[BCD]{\includegraphics[width=0.32\textwidth]{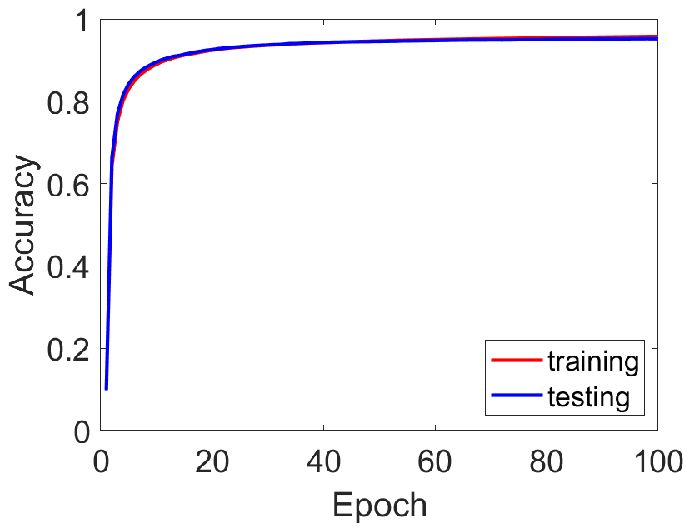}}
\subfigure[Proposed]{\includegraphics[width=0.32\textwidth]{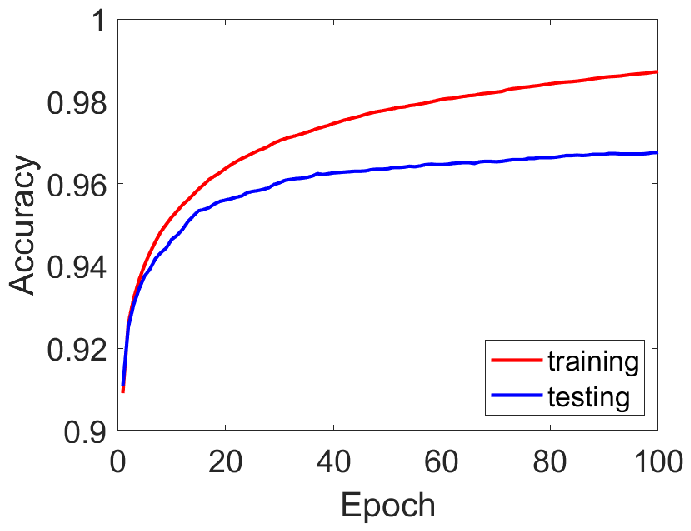}}
}
\caption{Comparison of average accuracy of (a) Adam, (b) BCD, and (c) proposed methods in classifying datasets from MNIST with learning rate of Adam optimizer set $0.001$, parameters of $BCD$ set at defaults, and initial settings of proposed method set according to Remark 2 in Algorithm \ref{alder-al}. Standard deviation of initial weights set at (top row) $0.1$, (middle row) $0.05$, and (bottom row) $0.01$. 
The accuracy of Adam and BCD is low at a standard deviation of $0.1$. The BCD does not smoothly increase with an increase in the number of iterations with standard deviation set at $0.05$. With standard deviation set at $0.01$, the results of all methods were acceptable.
}
\label{fig:mnist_class}
\end{center}
\end{figure}

\subsection{Compressed sensing recovery experiments}

We compared eight-layer ReLU networks (denoted as $\mathcal M_8$) derived using various learning methods in terms of compressed sensing recovery performance. The networks were learned using the Adam optimizer, BCD, and un-rectify methods using the same initial weights, the same number of epochs, and mini-batches of the same size.

\subsubsection{CS recovery using MNIST dataset}

Experiments were performed using the MNIST dataset. MNIST images ($28 \times 28$ in pixel) were  arranged as vectors ($784 \times 1$). Note that the average non-zero coefficient of images in MNIST is roughly $180$. The number of training data in MNIST is $60,000$ and the number of testing data in MNIST is $10,000$. Let the sensing matrix $\bA \in \R^{m \times n}$ be a Gaussian random matrix with elements sampled from i.i.d. Gaussian random variables with zero mean and variance $\frac{1}{m}$. Here, $n$ is set at $784$ and the value of $m$ varies ($10$, $25$, $100$, $200$, $300$, $400$, $500$, or $750$).
If the vector of MNIST is denoted as $\bx$, then $\by \in \R^{m \times 1}$ is the measurement obtained via $\by = \bA \bx$.
The MSE between $\bx$ and $\bar \bx$ is used as a performance metric in assessing the images derived using the un-rectify method (using the CGT-algorithmthat solves problem (PL)), the BCD method~\cite{zeng2018global}, and the back-propagation (BP) method with the Adam optimizer.
All of the networks learned using these methods had the same architecture (i.e., eight-layer ReLU networks). The sizes of each weight matrix $\bW$ and bias $\bb$ were $784\times784$ and $784\times1$, respectively. The number of learning epochs was set at $100$ and the mini-batch size was set at $6,000$.  The initial weights were obtained from i.i.d. Gaussian random variables with standard deviation of $0.01$, the value of which was derived from the results presented in Figures \ref{fig:mnist_simu},\ref{fig:mnist_err}, and \ref{fig:mnist_class}.
Figure \ref{fig:compare} compares reconstruction errors in terms of MSE versus the CS ratio for the recovery of MNIST testing data. All of the methods yielded satisfactory results; however, the proposed un-rectifying method was superior in all values of CS ratio. 
Figure \ref{fig:result1} compares the perceptual quality of the reconstructed images, with the original image displayed in the top row of Figure \ref{fig:mnist_simu}.
Consistent with the numerical results shown in Figure \ref{fig:compare}, the visual quality of the reconstructed images improved with an increase in the number of measurement. When the CS ratio fell  below $25\%$ (corresponding to $m = 200$), the Adam optimizer and the proposed un-rectifying methods achieved perceptual quality superior to that of the BCD. Nonetheless, at a CS ratio $25\%$ and higher, all three methods yielded images of consistent perceptual quality.

\begin{figure}[h]
\begin{center}
\includegraphics[width=0.9\textwidth]{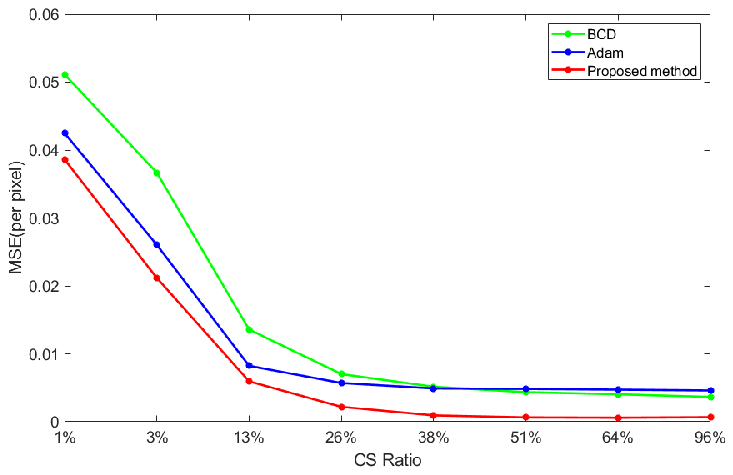}
  \caption{Comparison of average MSE (per pixel) from MNIST testing images versus CS ratio for  compressed sensing recovery problem with learning rate of Adam optimizer at $0.001$, parameters in BCD set at default values, and initial settings of proposed method set according to Remark 2 in Algorithm \ref{alder-al}. All MSE values decreased gradually as the CS ratio increased. 
    }\label{fig:compare}
\end{center}
\end{figure}

\begin{figure}[!ht]
\begin{center}
\mbox{
\fbox{\includegraphics[width=0.2\textwidth]{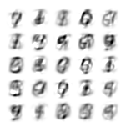}}
\fbox{\includegraphics[width=0.2\textwidth]{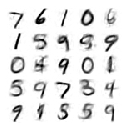}}
\fbox{\includegraphics[width=0.2\textwidth]{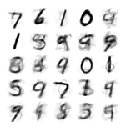}}
}
\mbox{
\fbox{\includegraphics[width=0.2\textwidth]{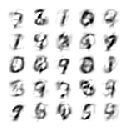}}
\fbox{\includegraphics[width=0.2\textwidth]{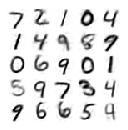}}
\fbox{\includegraphics[width=0.2\textwidth]{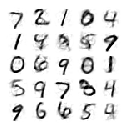}}
}
\mbox{
\fbox{\includegraphics[width=0.2\textwidth]{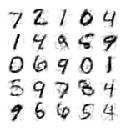}}
\fbox{\includegraphics[width=0.2\textwidth]{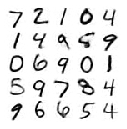}}
\fbox{\includegraphics[width=0.2\textwidth]{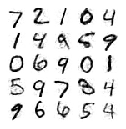}}
}
\mbox{
\subfigure[BCD]{\fbox{\includegraphics[width=0.2\textwidth]{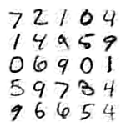}}}
\subfigure[Adam]{\fbox{\includegraphics[width=0.2\textwidth]{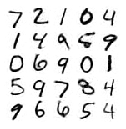}}}
\subfigure[Proposed]{\fbox{\includegraphics[width=0.2\textwidth]{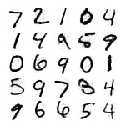}}}
}
\caption{Comparison of reconstructed image quality of original image in Figure \ref{fig:mnist_simu} provided by (a) BCD, (b) Adam, and (c) proposed methods with learning rate of Adam optimizer set at $0.001$, parameters in BCD set at default values, and initial settings of proposed method set according to Remark 2 in Algorithm \ref{alder-al}: $m$ values  of (first row) $10$, (second row) $25$, (third row) $100$, and (fourth row) $200$. }
\label{fig:result1}
\end{center}
\end{figure}

\subsubsection{CS recovery using natural images}
In this experiment, natural images were used for CS recovery. The $91$ images in \cite{zhang18ista} were used for training~\footnote{The images are publicly available from ``SRCNN-Tensorflow/Train at master · tegg89/SRCNN-Tensorflow-GitHub"}.  Patches ($32\times32$ pixels) were extracted from the luminance components of the images using a stride of $4$ in each dimension. The total number of training patches was $117,242$. The $11$ images in Set11 were used for testing. $58,523$ testing patches were obtained in a manner similar to that used in the extraction of training patches. $\bA \in \R^{m \times n}$ is a Gaussian random matrix with elements sampled from i.i.d. Gaussian random variables with zero mean and variance $\frac{1}{m}$. $n$ was set at $1024$ and the value of $m$ varied ($10$, $40$,  $102$, $256$, $409$, or $512$). Note that these values correspond to CS ratio of $1\%$, $4\%$, $10\%$, $25\%$, $40\%$, and $50\%$, respectively.
If the vector of the image patch is $\bx$ and the sensing matrix is $\bA$, then $\by \in \R^{m \times 1}$ is the measurement obtained via $\by = \bA \bx$. Let $\bX$ and $\bY$ denote the sets of training patches and their respective measurements.
The learning algorithms included in this comparison were the Adam optimizer, BCD, and proposed un-rectifying method. Eight-layer ReLU networks using $\bY$ as the input and $\bX$ as the desired output were learned. The sizes of the weight matrices $\bW_i$ and bias $\bb_i$ were set at $1024\times1024$ and $1024\times1$, respectively. The number of epochs was set at $200$. The mini-batch size was set at $512$. The initial weights used in all of the networks were sampled from i.i.d. Gaussian random variables with zero mean and standard deviation of $0.01$. The initial bias values of all affine mappings were set to zero. Reconstructed (testing) images were obtained by averaging the overlapped patches and then measuring the average peak-to-noise ratio (PSNR) and structural similarity index measure (SSIM). 
Tables \ref{psnr_compare} and  \ref{ssim_compare} respectively list the average PSNR and SSIM of images reconstructed using the three methods with various CS ratios.
The best performance for each CS ratio is highlighted in bold. The proposed un-rectifying method achieved the best performance in terms of PSNR and SSIM, regardless of CS ratio.
Figure \ref{fig:set11} presents images ($Barbara$, $House$, and $Parrot$) reconstructed with the CS ratio set at $10\%$ and $50\%$. The perceptual quality matches the results of Tables \ref{psnr_compare} and  \ref{ssim_compare}. Overall, the proposed method (CS ratio = $50 \%$) preserved more of the details, while providing shaper edges. The perceptual quality of the BCD and proposed methods were roughly the same at a CS ratio of $10\%$.

\begin{table}[h]
\begin{center}
\caption{Comparison of average PSNR(dB) in recovery of images in Set11 using varying CS ratios (best performances in bold).}
\small
\begin{tabular}{|l|llllll|}
\hline
\multirow{2}{*}{Method} & \multicolumn{6}{c|}{CS ratio} \\ \cline{2-7} 
   & 1\%   &   4\% &  10\%  &  25\% &  40\% & 50\% \\ \hline
Adam &  19.81  & 23.63 & 24.64 &  25.80  & 25.70& 25.89\\
BCD  &  19.51 & 22.86 & 25.93 &  29.02 & 26.89 & 26.87\\
Proposed & \textbf{19.93} &  \textbf{23.69} & \textbf{26.12}&\textbf{30.38} & \textbf{33.83} & \textbf{35.91}  \\ \hline
\end{tabular}\label{psnr_compare}
\end{center}
\end{table}

\begin{table}[h]
\begin{center}
\caption{Comparison of average SSIM in recovery of images in Set11 using various CS ratios (best performances in bold).}
\small
\begin{tabular}{|l|llllll|}
\hline
\multirow{2}{*}{Method} & \multicolumn{6}{c|}{CS ratio} \\ \cline{2-7} 
   & 1\%   &   4\% &  10\%  &  25\% &  40\% & 50\% \\ \hline
Adam &  0.55  & 0.71 & 0.74 &  0.79 & 0.78 & 0.79\\
BCD  &  0.54 & 0.69 & \textbf{0.82} &  0.90 &  0.89 & 0.85\\
Proposed & \textbf{0.56} &  \textbf{0.72}  & \textbf{0.82} &\textbf{0.92} & \textbf{0.96} & \textbf{0.97}  \\ \hline
\end{tabular}\label{ssim_compare}
\end{center}
\end{table}

\begin{figure}[h]
\begin{center}
\mbox{
{\begin{tikzpicture}[spy using outlines={rectangle,red,magnification=5,size=2.0cm}]
\node {\pgfimage[height=2cm]{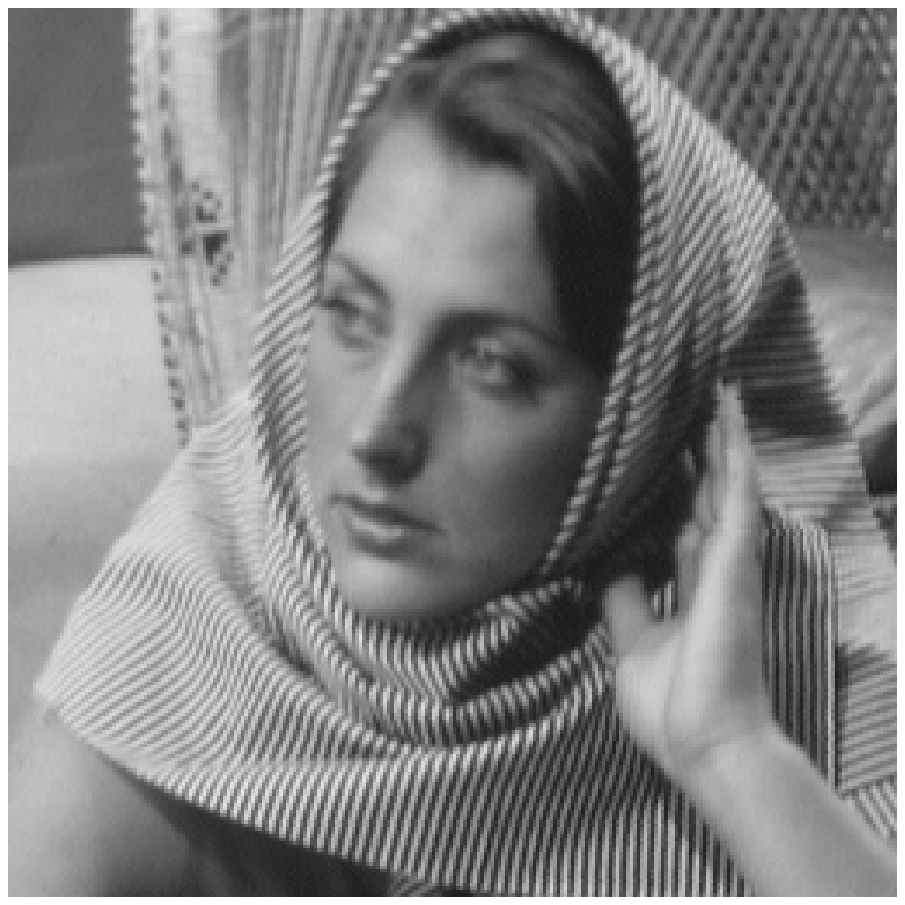}};
\spy on (0,-0.7) in node [left] at (1,-2.0);
\end{tikzpicture}}
{\begin{tikzpicture}[spy using outlines={rectangle,red,magnification=5,size=2.0cm}]
\node {\pgfimage[height=2cm]{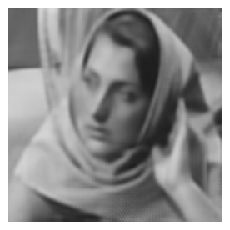}};
\spy on (0,-0.7) in node [left] at (1,-2.0);
\end{tikzpicture}}
{\begin{tikzpicture}[spy using outlines={rectangle,red,magnification=5,size=2.0cm}]
\node {\pgfimage[height=2cm]{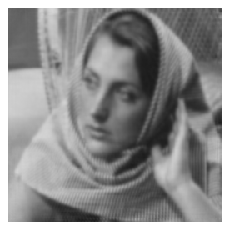}};
\spy on (0,-0.7) in node [left] at (1,-2.0);
\end{tikzpicture}}
{\begin{tikzpicture}[spy using outlines={rectangle,red,magnification=5,size=2.0cm}]
\node {\pgfimage[height=2cm]{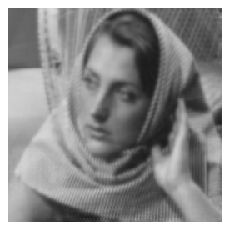}};
\spy on (0,-0.7) in node [left] at (1,-2.0);
\end{tikzpicture}}
{\begin{tikzpicture}[spy using outlines={rectangle,red,magnification=5,size=2.0cm}]
\node {\pgfimage[height=2cm]{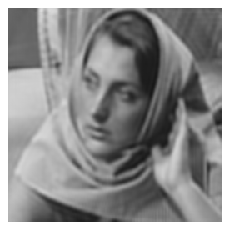}};
\spy on (0,-0.7) in node [left] at (1,-2.0);
\end{tikzpicture}}
{\begin{tikzpicture}[spy using outlines={rectangle,red,magnification=5,size=2.0cm}]
\node {\pgfimage[height=2cm]{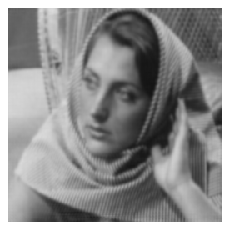}};
\spy on (0,-0.7) in node [left] at (1,-2.0);
\end{tikzpicture}}
{\begin{tikzpicture}[spy using outlines={rectangle,red,magnification=5,size=2.0cm}]
\node {\pgfimage[height=2cm]{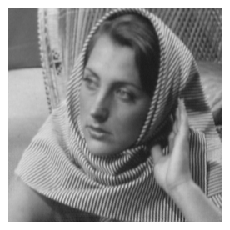}};
\spy on (0,-0.7) in node [left] at (1,-2.0);
\end{tikzpicture}}
}
\mbox{
{\begin{tikzpicture}[spy using outlines={rectangle,green,magnification=5,size=2.0cm}]
\node {\pgfimage[height=2cm]{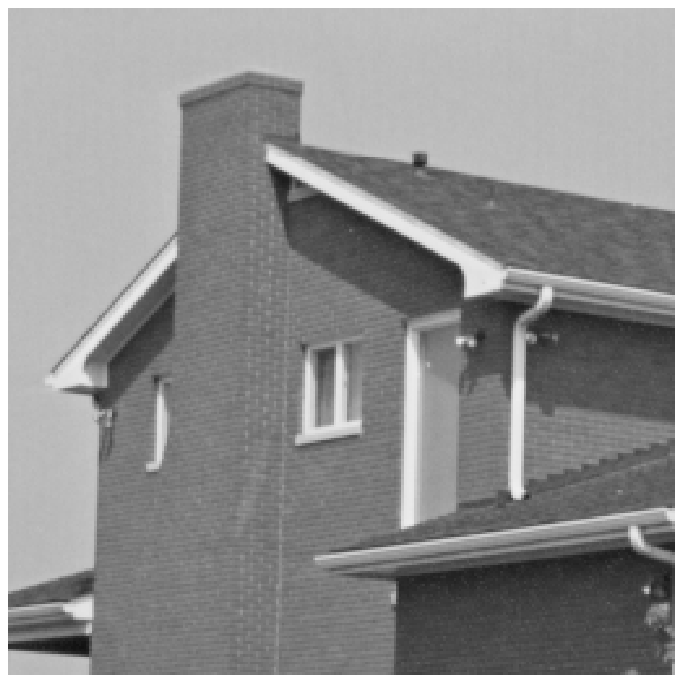}};
\spy on (-0.7,-0.1) in node [left] at (1,-2.0);
\end{tikzpicture}}
{\begin{tikzpicture}[spy using outlines={rectangle,green,magnification=5,size=2.0cm}]
\node {\pgfimage[height=2cm]{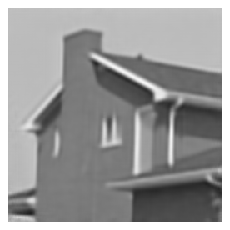}};
\spy on (-0.7,-0.1) in node [left] at (1,-2.0);
\end{tikzpicture}}
{\begin{tikzpicture}[spy using outlines={rectangle,green,magnification=5,size=2.0cm}]
\node {\pgfimage[height=2cm]{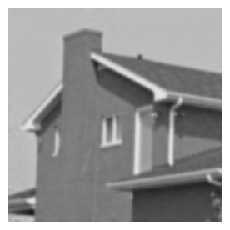}};
\spy on (-0.7,-0.1) in node [left] at (1,-2.0);
\end{tikzpicture}}
{\begin{tikzpicture}[spy using outlines={rectangle,green,magnification=5,size=2.0cm}]
\node {\pgfimage[height=2cm]{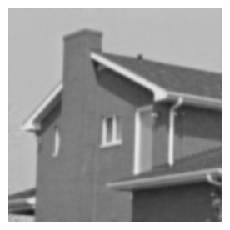}};
\spy on (-0.7,-0.1) in node [left] at (1,-2.0);
\end{tikzpicture}}
{\begin{tikzpicture}[spy using outlines={rectangle,green,magnification=5,size=2.0cm}]
\node {\pgfimage[height=2cm]{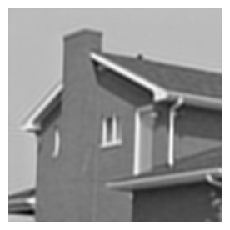}};
\spy on (-0.7,-0.1) in node [left] at (1,-2.0);
\end{tikzpicture}}
{\begin{tikzpicture}[spy using outlines={rectangle,green,magnification=5,size=2.0cm}]
\node {\pgfimage[height=2cm]{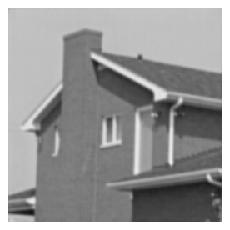}};
\spy on (-0.7,-0.1) in node [left] at (1,-2.0);
\end{tikzpicture}}
{\begin{tikzpicture}[spy using outlines={rectangle,green,magnification=5,size=2.0cm}]
\node {\pgfimage[height=2cm]{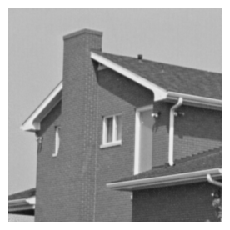}};
\spy on (-0.7,-0.1) in node [left] at (1,-2.0);
\end{tikzpicture}}
}
\mbox{
\subfigure[original]{\begin{tikzpicture}[spy using outlines={rectangle,blue,magnification=5,size=2.0cm}]
\node {\pgfimage[height=2cm]{parrots.eps}};
\spy on (0.2,0.3) in node [left] at (1,-2.0);
\end{tikzpicture}}
\subfigure[Adam $10\%$]{\begin{tikzpicture}[spy using outlines={rectangle,blue,magnification=5,size=2.0cm}]
\node {\pgfimage[height=2cm]{bp10_parrots.eps}};
\spy on (0.2,0.3) in node [left] at (1,-2.0);
\end{tikzpicture}}
\subfigure[BCD $10\%$]{\begin{tikzpicture}[spy using outlines={rectangle,blue,magnification=5,size=2.0cm}]
\node {\pgfimage[height=2cm]{bcd10_parrots.eps}};
\spy on (0.2,0.3) in node [left] at (1,-2.0);
\end{tikzpicture}}
\subfigure[Our $10\%$]{\begin{tikzpicture}[spy using outlines={rectangle,blue,magnification=5,size=2.0cm}]
\node {\pgfimage[height=2cm]{our10_parrots.eps}};
\spy on (0.2,0.3) in node [left] at (1,-2.0);
\end{tikzpicture}}
\subfigure[Adam $50\%$]{\begin{tikzpicture}[spy using outlines={rectangle,blue,magnification=5,size=2.0cm}]
\node {\pgfimage[height=2cm]{bp50_parrots.eps}};
\spy on (0.2,0.3) in node [left] at (1,-2.0);
\end{tikzpicture}}
\subfigure[BCD $50\%$]{\begin{tikzpicture}[spy using outlines={rectangle,blue,magnification=5,size=2.0cm}]
\node {\pgfimage[height=2cm]{bcd50_parrots.eps}};
\spy on (0.2,0.3) in node [left] at (1,-2.0);
\end{tikzpicture}}
\subfigure[Our $50\%$]{\begin{tikzpicture}[spy using outlines={rectangle,blue,magnification=5,size=2.0cm}]
\node {\pgfimage[height=2cm]{our50_parrots.eps}};
\spy on (0.2,0.3) in node [left] at (1,-2.0);
\end{tikzpicture}}
}
\caption{Comparison of perceptual quality of reconstructed images in Set11 dataset wiht CS ratio set at $10\%$ and $50\%$ with learning rate of Adam optimizer set at $0.001$ and parameters in BCD set at default values: (top) $Barbara$; (middle) $House$; and (bottom) $Parrot$ (insets show magnification of highlighted area). }
\label{fig:set11}
\end{center}
\end{figure}

\section{Conclusions} \label{conclusions}

This paper demonstrates that the deep learning problem with a twice-continuously-differentiable objective function can be solved using the augmented Lagrangian approach, based on the algorithm developed by Conn, Could, and Toint. Note that this conclusion cannot be derived directly, due to the fact that ReLUs are not  differentiable functions. Un-rectifying a ReLU network makes it possible to formulate a learning problem as a twice-continuously-differentiable non-convex optimization problem with twice-continuously-differentiable data-dependent constraints . For this problem, activation variables corresponding to un-rectifying ReLUs are relaxed to data-dependent variables within domains confined to closed segments. 
We demonstrate that un-rectifying networks can converge to local optima when the updating of primal variables is subjected to alternating minimization. We compared the performance of the proposed scheme  with other state-of-the-art deep learning methods commonly applied to compressed sensing recovery. 
In experiments, the proposed scheme demonstrated superior robustness to the initial weight matrices and achieved performance superior to that of other state-of-the-art methods. In this paper, we applied the proposed optimization method to learning ReLU networks; however, this method could theoretically be applied to any network comprising continuous, point-wise, piecewise linear activation functions. Our theoretical results assume a batch input.  The question of whether theoretical results could be derived using mini-batch inputs is a subject worthy of further study. Applying the un-rectifying technique to problems of classification may be another worthwhile avenue of inquiry.

\bibliographystyle{ieeetr}
\bibliography{unrectify}

\begin{thebibliography}{10}

\bibitem{balestriero2018mad}
R.~Balestriero and R.~Baraniuk, ``Mad max: Affine spline insights into deep
  learning,'' {\em arXiv preprint arXiv:1805.06576}, 2018.

\bibitem{Wen19}
W.~L. Hwang and A.~Heinecke, ``Un-rectifying non-linear networks for signal
  representation,'' {\em IEEE Trans. on Signal Processing (Accepted)}.

\bibitem{gong2014compressing}
Y.~Gong, L.~Liu, M.~Yang, and L.~Bourdev, ``Compressing deep convolutional
  networks using vector quantization,'' {\em arXiv preprint arXiv:1412.6115},
  2014.

\bibitem{he2016deep}
K.~He, X.~Zhang, S.~Ren, and J.~Sun, ``Deep residual learning for image
  recognition,'' in {\em Proceedings of the IEEE conference on computer vision
  and pattern recognition}, pp.~770--778, 2016.

\bibitem{gregor2010learning}
K.~Gregor and Y.~LeCun, ``Learning fast approximations of sparse coding,'' in
  {\em Proceedings of the 27th international conference on international
  conference on machine learning}, pp.~399--406, 2010.

\bibitem{kuo2018data}
C.-C.~J. Kuo and Y.~Chen, ``On data-driven saak transform,'' {\em Journal of
  Visual Communication and Image Representation}, vol.~50, pp.~237--246, 2018.

\bibitem{szegedy2016inception}
C.~Szegedy, S.~Ioffe, V.~Vanhoucke, and A.~Alemi, ``Inception-v4,
  inception-resnet and the impact of residual connections on learning,'' {\em
  arXiv preprint arXiv:1602.07261}, 2016.

\bibitem{kung2020term}
H.~Kung, B.~McDanel, and S.~Q. Zhang, ``Term quantization: furthering
  quantization at run time,'' in {\em Proceedings of the International
  Conference for High Performance Computing, Networking, Storage and Analysis},
  pp.~1--14, 2020.

\bibitem{huang2017densely}
G.~Huang, Z.~Liu, L.~Van Der~Maaten, and K.~Q. Weinberger, ``Densely connected
  convolutional networks,'' in {\em Proceedings of the IEEE conference on
  computer vision and pattern recognition}, pp.~4700--4708, 2017.

\bibitem{zhang2018ista}
J.~Zhang and B.~Ghanem, ``Ista-net: Interpretable optimization-inspired deep
  network for image compressive sensing,'' in {\em Proceedings of the IEEE
  conference on computer vision and pattern recognition}, pp.~1828--1837, 2018.

\bibitem{zeng2018global}
J.~Zeng, T.~T.-K. Lau, S.~Lin, and Y.~Yao, ``Global convergence of block
  coordinate descent in deep learning,'' {\em arXiv preprint arXiv:1803.00225},
  2018.

\bibitem{gao2020admm}
W.~Gao, D.~Goldfarb, and F.~E. Curtis, ``Admm for multiaffine constrained
  optimization,'' {\em Optimization Methods and Software}, vol.~35, no.~2,
  pp.~257--303, 2020.

\bibitem{Ara18}
R.~Arora, A.~Basu, P.~Mianjy, and A.~Mukherjee, ``Understanding deep neural
  networks with rectified linear units,'' in {\em International Conference on
  Learning Representations (ICLR)}, 2018.

\bibitem{serra2017bounding}
T.~Serra, C.~Tjandraatmadja, and S.~Ramalingam, ``Bounding and counting linear
  regions of deep neural networks,'' {\em arXiv preprint arXiv:1711.02114},
  2017.

\bibitem{HintonReLU}
V.~Nair and G.~E. Hinton, ``Rectified linear units improve restricted
  {B}oltzmann machines,'' in {\em Proceedings of the 27th International
  Conference on Machine Learning (ICML)}, pp.~807--814, 2010.

\bibitem{GlorotBordesBengio2011}
X.~Glorot, A.~Bordes, and Y.~Bengio, ``Deep sparse rectifier neural networks,''
  in {\em Proceedings of the 14th International Conference on Artificial
  Intelligence and Statistics}, vol.~15, pp.~315--323, PMLR, 2011.

\bibitem{heinecke2020refinement}
A.~Heinecke, J.~Ho, and W.-L. Hwang, ``Refinement and universal approximation
  via sparsely connected relu convolution nets,'' {\em IEEE Signal Processing
  Letters}, vol.~27, pp.~1175--1179, 2020.

\bibitem{conn1991globally}
A.~R. Conn, N.~I. Gould, and P.~Toint, ``A globally convergent augmented
  lagrangian algorithm for optimization with general constraints and simple
  bounds,'' {\em SIAM Journal on Numerical Analysis}, vol.~28, no.~2,
  pp.~545--572, 1991.

\bibitem{kingma2014adam}
D.~P. Kingma and J.~Ba, ``Adam: A method for stochastic optimization,'' {\em
  arXiv preprint arXiv:1412.6980}, 2014.

\bibitem{taylor2016training}
G.~Taylor, R.~Burmeister, Z.~Xu, B.~Singh, A.~Patel, and T.~Goldstein,
  ``Training neural networks without gradients: A scalable admm approach,'' in
  {\em International conference on machine learning (ICML)}, pp.~2722--2731,
  2016.

\bibitem{lau2018proximal}
T.~T.-K. Lau, J.~Zeng, B.~Wu, and Y.~Yao, ``A proximal block coordinate descent
  algorithm for deep neural network training,'' {\em arXiv preprint
  arXiv:1803.09082}, 2018.

\bibitem{rumelhart1986learning}
D.~E. Rumelhart, G.~E. Hinton, and R.~J. Williams, ``Learning representations
  by back-propagating errors,'' {\em nature}, vol.~323, no.~6088, pp.~533--536,
  1986.

\bibitem{glorot10}
X.~Glorot and Y.~Bengio, ``Understanding the difficulty of training deep
  feedforward neural networks,'' in {\em Proceedings of Machine Learning
  Research}, pp.~249--256, 2010.

\bibitem{he2015delving}
K.~He, X.~Zhang, S.~Ren, and J.~Sun, ``Delving deep into rectifiers: Surpassing
  human-level performance on imagenet classification,'' in {\em Proceedings of
  the IEEE international conference on computer vision}, pp.~1026--1034, 2015.

\bibitem{krizhevsky2017imagenet}
A.~Krizhevsky, I.~Sutskever, and G.~E. Hinton, ``Imagenet classification with
  deep convolutional neural networks,'' {\em Communications of the ACM},
  vol.~60, no.~6, pp.~84--90, 2017.

\bibitem{ioffe2015batch}
S.~Ioffe and C.~Szegedy, ``Batch normalization: Accelerating deep network
  training by reducing internal covariate shift,'' {\em arXiv preprint
  arXiv:1502.03167}, 2015.

\bibitem{nesterov1983method}
Y.~Nesterov, ``A method for unconstrained convex minimization problem with the
  rate of convergence o (1/k\^{} 2),'' in {\em Doklady an ussr}, vol.~269,
  pp.~543--547, 1983.

\bibitem{duchi2011adaptive}
J.~Duchi, E.~Hazan, and Y.~Singer, ``Adaptive subgradient methods for online
  learning and stochastic optimization.,'' {\em Journal of machine learning
  research}, vol.~12, no.~7, 2011.

\bibitem{ruder2016overview}
S.~Ruder, ``An overview of gradient descent optimization algorithms,'' {\em
  arXiv preprint arXiv:1609.04747}, 2016.

\bibitem{robbins1951stochastic}
H.~Robbins and S.~Monro, ``A stochastic approximation method,'' {\em The annals
  of mathematical statistics}, pp.~400--407, 1951.

\bibitem{bottou2018optimization}
L.~Bottou, F.~E. Curtis, and J.~Nocedal, ``Optimization methods for large-scale
  machine learning,'' {\em Siam Review}, vol.~60, no.~2, pp.~223--311, 2018.

\bibitem{zhang2017convergent}
Z.~Zhang and M.~Brand, ``Convergent block coordinate descent for training
  tikhonov regularized deep neural networks,'' in {\em Advances in Neural
  Information Processing Systems (NIPS)}, pp.~1721--1730, 2017.

\bibitem{gu2018fenchel}
F.~Gu, A.~Askari, and L.~E. Ghaoui, ``Fenchel lifted networks: A lagrange
  relaxation of neural network training,'' {\em International conference on
  machine learning (ICML)}, 2018.

\bibitem{kurdyka1998gradients}
K.~Kurdyka, ``On gradients of functions definable in o-minimal structures,'' in
  {\em Annales de l'institut Fourier}, vol.~48, pp.~769--783, 1998.

\bibitem{lojasiewicz1993geometrie}
S.~{\L}ojasiewicz, ``Sur la g{\'e}om{\'e}trie semi-et sous-analytique,'' in
  {\em Annales de l'institut Fourier}, vol.~43, pp.~1575--1595, 1993.

\bibitem{attouch2013convergence}
H.~Attouch, J.~Bolte, and B.~F. Svaiter, ``Convergence of descent methods for
  semi-algebraic and tame problems: proximal algorithms, forward--backward
  splitting, and regularized gauss--seidel methods,'' {\em Mathematical
  Programming}, vol.~137, no.~1-2, pp.~91--129, 2013.

\bibitem{xu2017globally}
Y.~Xu and W.~Yin, ``A globally convergent algorithm for nonconvex optimization
  based on block coordinate update,'' {\em Journal of Scientific Computing},
  vol.~72, no.~2, pp.~700--734, 2017.

\bibitem{bertsekas2014constrained}
D.~P. Bertsekas, {\em Constrained optimization and Lagrange multiplier
  methods}.
\newblock Academic press, 2014.

\bibitem{choromanska2018beyond}
A.~Choromanska, B.~Cowen, S.~Kumaravel, R.~Luss, M.~Rigotti, I.~Rish,
  B.~Kingsbury, P.~DiAchille, V.~Gurev, R.~Tejwani, {\em et~al.}, ``Beyond
  backprop: Online alternating minimization with auxiliary variables,'' {\em
  arXiv preprint arXiv:1806.09077}, 2018.

\bibitem{glowinski1975approximation}
R.~Glowinski and A.~Marroco, ``Sur l'approximation, par {\'e}l{\'e}ments finis
  d'ordre un, et la r{\'e}solution, par p{\'e}nalisation-dualit{\'e} d'une
  classe de probl{\`e}mes de dirichlet non lin{\'e}aires,'' {\em ESAIM:
  Mathematical Modelling and Numerical Analysis-Mod{\'e}lisation
  Math{\'e}matique et Analyse Num{\'e}rique}, vol.~9, no.~R2, pp.~41--76, 1975.

\bibitem{eckstein2015understanding}
J.~Eckstein and W.~Yao, ``Understanding the convergence of the alternating
  direction method of multipliers: Theoretical and computational
  perspectives,'' {\em Pac. J. Optim.}, vol.~11, no.~4, pp.~619--644, 2015.

\bibitem{magnusson2015convergence}
S.~Magn{\'u}sson, P.~C. Weeraddana, M.~G. Rabbat, and C.~Fischione, ``On the
  convergence of alternating direction lagrangian methods for nonconvex
  structured optimization problems,'' {\em IEEE Transactions on Control of
  Network Systems}, vol.~3, no.~3, pp.~296--309, 2015.

\bibitem{wang2019global}
Y.~Wang, W.~Yin, and J.~Zeng, ``Global convergence of admm in nonconvex
  nonsmooth optimization,'' {\em Journal of Scientific Computing}, vol.~78,
  no.~1, pp.~29--63, 2019.

\bibitem{zeng2019convergence}
J.~Zeng, S.-B. Lin, and Y.~Yao, ``A convergence analysis of nonlinearly
  constrained admm in deep learning,'' {\em arXiv preprint arXiv:1902.02060},
  2019.

\bibitem{zeng2020admm}
J.~Zeng, S.-B. Lin, Y.~Yao, and D.-X. Zhou, ``On admm in deep learning:
  Convergence and saturation-avoidance,'' {\em arXiv preprint
  arXiv:1902.02060}, 2020.

\bibitem{BaraniukPowerDiagramSubdiv}
R.~Balestriero, R.~Cosentino, B.~Aazhang, and R.~Baraniuk, ``The geometry of
  deep networks: Power diagram subdivision,'' {\em Advances Neural Inf.
  Process. Syst.}, pp.~15806--15815, 2019.

\bibitem{Ber08}
D.~P. Bertsekas, ``Nonlinear programming: 3rd,'' {\em Athena Scientific
  Optimization and Computations Series 4}, vol.~4, 2008.

\bibitem{mousavi2017learning}
A.~Mousavi and R.~G. Baraniuk, ``Learning to invert: Signal recovery via deep
  convolutional networks,'' in {\em 2017 IEEE international conference on
  acoustics, speech and signal processing (ICASSP)}, pp.~2272--2276, IEEE,
  2017.

\bibitem{he15}
K.~{He}, X.~{Zhang}, S.~{Ren}, and J.~{Sun}, ``Delving deep into rectifiers:
  Surpassing human-level performance on imagenet classification,'' in {\em IEEE
  International Conference on Computer Vision}, pp.~1026--1034, 2015.

\bibitem{zhang18ista}
J.~Zhang and B.~Ghanem, ``Ista-net: Interpretable optimization-inspired deep
  network for image compressive sensing,'' in {\em IEEE Conference on Computer
  Vision and Pattern Recognition}, pp.~1828--1837, 2018.

\end{thebibliography}

\appendix
\section{Proof of Theorem \ref{OURmaincriticalpoint} }\label{sec:globalconv}

This appendix outlines the method used to obtain the conclusion for Theorem \ref{OURmaincriticalpoint}. The following analysis uses many symbols that are self-contained. Despite efforts to avoid overloading the manuscript with notation, a certain number was unavoidable.

Recall that we consider the problem of finding a local minimizer of the function 
\begin{align} \label{P11}
f(\bx^1,\bx^2) \colon \R^n \to \R, \text{ where $\bx^1 \in \R^{n_1}$ and $\bx^2 \in\R^{n_2}$}
\end{align}
where $n = n_1 + n_2$, $\bx^1$ and $(\bx^1, \bx^2)$ are required to satisfy
\begin{align} \label{P21}
c_j(\bx^1, \bx^2) = 0,  \text{ $1 \leq j \leq m$,}
\end{align}
where there is no constraint on $\bx^1$, and any component $x_i$ of $\bx^2$ is required to satisfy the simple bounds for $u_i > l_i$ where $l_i \in \R$ and $u_i \in \R \cup \{\infty\}$ and 
\begin{align} \label{P3}
l_i \leq x_i \leq u_i.
\end{align}
Note that $B$ denote the convex set of $\bx = (\bx^1, \bx^2)$ wherein $\bx^2$ satisfies (\ref{P3}).

\subsection{Machinery and notation}

As defined in Section \ref{sec:globalconv1}, we assume that (AS1)-(AS3) hold in the following derivations. 
We define the component-wise projection operator
\begin{align*} 
(\bP[\bx])_i = \begin{cases}
l_i \text{ if $x_i$ is a component of $\bx^2$ and $x_i \leq l_i$}, \\
u_i \text{ if $x_i$ is a component of $\bx^2$ and $x_i \geq u_i$}, \\
x_i \text{ otherwise}
\end{cases}
\end{align*}
and
\begin{align} \label{projection}
& (\bP(\bx, \bv) )_i =   (\bx - \bP[\bx-\bv])_i  \nonumber \\
& = 
\begin{cases}
v_i & \text{ if $x_i$ is a component of $\bx^1$}, \\
v_i & \text{ if $l_i < x_i - v_i < u_i$, $x_i$ is a component of $\bx^2$}, \\
x_i - l_i & \text{ if $x_i - v_i \leq l_i$, $x_i$ is a a component of $\bx^2$,} \\
x_i - u_i & \text{ if $x_i - v_i \geq u_i$, $x_i$ is a a component of $\bx^2$}. 
\end{cases}
\end{align}
To clarify (\ref{projection}), we provide the following intuitive interpretation, which can be related to the variational inequality in situations where $\bP(\bx, \bv) = 0$: $\bv$ can be regarded as the gradient of a continuous function, such that $\bx - \bv$ can be regarded as the next iterator. 
For any sub-sequence of $(\bP(\bx^{(k)}, \bv^{(k)}))_i$ (i.e., the $i$-th sub-component) convergent to zero, the corresponding sub-sequence of $x^{(k)}_i$ is convergent to either a boundary point ($l_i$ or $u_i$) or an interior point of $B$, where the $i$-th sub-component of the gradient at the point is zero.

The Lagrangian function and augmented Lagrangian function of problem (\ref{P11})-(\ref{P3}) with respect to constraints $c_i$ (where $\lambda_i$ are Lagrangian multipliers) can respectively be written as follows:
\begin{align} \label{lagrangian}
L(\bx, \lambda) = f(\bx) + \sum_{j=1}^m \lambda_j c_j(\bx),
\end{align}
and 
\begin{align} \label{appxauglag}
L_{\mu}(\bx, \lambda) = f(\bx) + \sum_{j=1}^m \lambda_j c_j(\bx) + \frac{1}{2\mu} \sum_{j=1}^m c_j(\bx)^2.
\end{align}
Note that we do not include the simple bounds (\ref{P3}) in the augmented Lagrangian function. Instead, we seek to ensure that these constraints are always satisfied in the sequential minimization of primal variables. 
If we denote $\lambda = [\lambda_1, \ldots, \lambda_m]^\top$, $c(\bx) =  [c_1(\bx), \ldots, c_m(\bx)]^\top$, and let 
\begin{align} \label{barlambda}
\bar \lambda = \lambda + c(\bx)/\mu,
\end{align}
then 
\begin{align} \label{auglanandlan}
\nabla_{\bx} L_{\mu}(\bx, \lambda) =\nabla_{\bx} L(\bx, \bar \lambda) = \nabla_{\bx} f(\bx) + \sum_{j=1}^m \bar \lambda_j \nabla_{\bx} c_j(\bx).
\end{align}

Let the iterators $\{\bx^{(k)}\in B\}$, in which $l_i \leq x_i^{(k)} \leq u_i$ hold for $x_i \in \bx^2$, and let $\{\lambda^{(k)} \in \R^m\}$ be the sequence of Lagrangian multipliers and $\{ \mu^{(k)} > 0\}$ be a sequence of positive scalars. For any function $F$ in this section, we use notation $F^{(k)}$ to denote $F$ evaluated using arguments $\bx^{(k)}, \lambda^{(k)}$, or $\mu^{(k)}$ as appropriate. For example, 
\begin{align} \label{gradientofL}
\nabla_{\bx}L_{\mu}^{(k)}= \nabla_{\bx} L_{\mu^{(k)}}(\bx^{(k)}, \lambda^{(k)}). 
\end{align}
Consider the variational inequality for an optimal solution to augmented Lagrangian function (\ref{appxauglag}), which is as follows:
\begin{align}
\begin{cases}
(\nabla_{\bx} L_{\mu}(\bx^*, \lambda^*))_i   = 0 \text{ if $x_i$ is a component in $\bx^1$; otherwise,} \\
\begin{cases}
(\nabla_{\bx} L_{\mu}(\bx^*, \lambda^*))_i (x_i - l_i) \geq 0 \text{ or } \\
(\nabla_{\bx} L_{\mu}(\bx^*, \lambda^*))_i (x_i - u_i) \geq 0 \text{ for $x_i \in [l_i, u_i]$,}
\end{cases}
\end{cases}
\end{align}
Then, we have three possibilities for any component $x_i$ of $\bx^2$: for $\bx^{(k)}$, $\lambda^{(k)}$, and $\mu^{(k)}$, 
\begin{align}
\text{(i) }& 0 \leq x_i^{(k)}  - l_i \leq (\nabla_{\bx}L_{\mu}^{(k)})_i, \label{case1} \\
\text{(ii) }& (\nabla_{\bx}L_{\mu}^{(k)})_i \leq x_i^{(k)} - u_i \leq 0,\text{ or }  \label{case2}\\
\text{(iii) }&  x_i^{(k)} - u_i < (\nabla_{\bx}L_{\mu}^{(k)})_i < x_i^{(k)} - l_i. \label{case3}
\end{align}

For any component $x_i$ of $\bx^2$ in case (i), we apply (\ref{projection}) to obtain
\begin{align} \label{dominatedabove}
(\bP(\bx^{(k)}, \nabla_{\bx}L_{\mu}^{(k)} )_{n_1+i} = x_i^{(k)} - l_i.
\end{align}
In case (ii), we obtain  
\begin{align} \label{dominatedbelow}
(\bP(\bx^{(k)}, \nabla_{\bx}L_{\mu}^{(k)})_{n_1+i} = x_i^{(k)} - u_i,
\end{align}
and in case (iii), we obtain 
\begin{align}
(\bP(\bx^{(k)}, \nabla_{\bx}L_{\mu}^{(k)} )_{n_1+i} = (\nabla_{\bx}L_{\mu}^{(k)})_{n_1+i}. \label{floating2}
\end{align}

For any component $x_i$ of $\bx^1$ we apply (\ref{projection}) to obtain
\begin{align} \label{floating1}
(\bP(\bx^{(k)},\nabla_{\bx}L_{\mu}^{(k)})_i = (\nabla_{\bx}L_{\mu}^{(k)})_i. 
\end{align}

Furthermore, if there is a convergent sequence $\{\bx^{(k)}, k \in K\}$, with limit point $\bx^*$, then for any component $x_i$ of $\bx$, $x_i^{(k)}$, all $k \in K$ of sufficiently large can be partitioned into the following sets related to the three possibilities described above as well as the corresponding $\bx^*$, as follows:
{\small
\begin{align}
\begin{cases}
I_0 = \{i | x_i^{(k)}\text{ meets } (\ref{floating1}) \text{ where $x_i$  is a component of $\bx^1$}\}, \\
I_1 = \{i| \text{$x_i^{(k)}$ meets (iii) and (\ref{floating2})} \\ \text{ with $x_i^*$ is in the interior of $B$ 
where $x_i$ is a component of $\bx^2$}\},\\
I_{2l} = \{i|\text{$x_i^{(k)}$ meets (i) and (\ref{dominatedabove});  $x_i$ is a component of $\bx^2$} \},\\
I_{2u} = \{i|\text{$x_i^{(k)}$ meets (ii) and (\ref{dominatedbelow}) where $x_i$ is a component of $\bx^2$} \}, \\
I_{3l} = \{i| \text{$x_i^{(k)}$ meets (iii) and (\ref{floating2}) } \\ \text{ with $x_i^* = l_i$ where $x_i$ s a component of $\bx^2$} \}, \\
I_{3u} = \{i| \text{$x_i^{(k)}$ meets (iii) and (\ref{floating2}) } \\ \text{ with $x_i^* = u_i$ where $x_i$ is a component of $\bx^2$} \}, \\
I_4 = \text{ indices of components of $\bx^2$ others than any of the above.}
\end{cases}
\end{align}}
In the above, $I_0$ denotes the indices of variables in $\bx^1$.  
$I_1$, $I_{3l}$, and $I_{3u}$ comprise the indices of variables in $\bx^2$ meeting (iii) and respectively converge to the interior of $B$, the lower bound, and the upper bound.  Occasionally, for the sake of convenience, we abuse the notation by referring to elements in $I_i$ as variables rather than indices of variables. For instance, $I_{2l}$ and $I_{2u}$ are made up of variables in $\bx^2$ that respectively meet (i) and (ii), when $k$ is sufficiently large. The following result is derived by selecting the iterates in a manner where $\bP(\bx^{(k)}, \nabla_{\bx}L_{\mu}^{(k)})$ approaches zero with an increase in $k \in K$.

\begin{citedlem} \label{glolem1}
Suppose that $\{\bx^{(k)}, k \in K\}$ is a convergent sequence with limit point $\bx^*$ and $\bP(\bx^{(k)}, \nabla_{\bx}L_{\mu}^{(k)})$ approaches zero with an increase in $k \in K$.  \\
(i) The variables (with indices) in $I_{2l}$, $I_{2u}$, $I_{3l}$, and $I_{3u}$ all converge to their bounds.\\
(ii) The components of $\nabla_{\bx}L_{\mu}^{(k)}$ (with indices) in $I_0$, $I_1$, $I_{3l}$, and $I_{3u}$ converge to zero.\\
(iii) There must be a subsequence of $I_4$ that converges to one of their bounds. If a component of $\nabla_{\bx} L_{\mu}^{(k)}$ (with index) in $I_4$ converges to a finite limit, then the limit is zero.
\end{citedlem}
\proof 
(i) The result is true for variables in $I_{2l}$ and $I_{2u}$ as respectively derived from (\ref{dominatedabove}) and (\ref{dominatedbelow}). By definition, it is true for variables in $I_{3l}$ and $I_{3u}$. \\
(ii) The result for variables in $I_0$ follows from (\ref{floating1}) and is therefore true.  The results for variables in $I_1$, $I_{3l}$, and $I_{3u}$ follow from (\ref{floating2}). \\
(iii) 
For variables in $I_4$, from (\ref{dominatedabove}), (\ref{dominatedbelow}),(\ref{floating2}), and $u_i > l_i$, there must be a subsequence of $I_4$ that converges to either $u_i$ or $l_i$.  The assertion is obviously true for a subsequence converging to an interior point of the interval. Thus, we suppose that $x^*_i = l_i$,
and the sub-sequence converging to the other bound can be derived in a similar manner.  There must be an infinite subsequence of $k \in K$ where (\ref{case3}) holds. Following from (\ref{floating2}) and the fact that $\bP(\bx^{(k)}, \nabla_{\bx}L_{\mu}^{(k)})$ approaches zero, it
would follow that $\nabla_{\bx} L_{\mu}^{(k)}$ converges to zero with an increase in $k \in K$. 
\qed

Recall that $\bA(\bx)$ denotes the $m \times n$ Jacobian of $c(\bx) =  [c_1(\bx), \ldots, c_m(\bx)]^\top$ and $J = I_0 \cup I_1$ ($x_i$ is in $\bx^1$ or is in $\bx^2$ with the value strictly in the interior of closed interval $[l_i, u_i]$). Let $\tilde n$ denote the size of $J$. We use notation $\bA(\bx)|_J$ to denote the $m \times \tilde n$ sub-matrix of $\bA(\bx)$ with columns corresponding to variables (with indices) in $J$ and use notation $\nabla f(\bx)|_J \in \R^{\tilde n}$ to denote the sub-vector of $\nabla f(\bx)$ with components indexed in $J$.

For $\bx \in B$, we can obtain the least-square Lagrangian multiplier estimate $\hat \lambda(\bx)$ by taking the partial derivative of Lagrangian function $L$ (i.e.,  (\ref{lagrangian})) with respect to variables in $J$ and then set the result to zero to derive the following estimate of the Lagrangian multiplier:
\begin{align} \label{lambdaestimate}
\hat \lambda(\bx) = - ((\bA(\bx)|_J)^+)^\top \nabla f(\bx)|_J
\end{align}
where 
\begin{align*}
(\bA(\bx)|_J)^+ = (\bA(\bx)|_J)^\top [\bA(\bx)|_J(\bA(\bx)|_J)^\top]^{-1}.
\end{align*}

\subsection{Analysis of convergence}

The following lemma demonstrates that the CGT-algorithm deals with the sequences of Lagrangian multiplier $\lambda^{(k)}$ in a manner where $\mu^{(k)} \|\lambda^{(k)}\| \rightarrow 0$ when $k \rightarrow \infty$.

\begin{citedlem}(Lemma 4.2 in  \cite{conn1991globally})
Suppose that $\mu^{(k)} \rightarrow 0$ when the CGT-algorithm is executed. Then the product $\mu^{(k)} \|\lambda^{(k)}\| \rightarrow 0$.
\end{citedlem}

The following two lemmas are required to support the proof of Theorem \ref{OURmaincriticalpoint}.

\begin{citedlem} (Lemma 4.3 in  \cite{conn1991globally})\label{glolem2}
Suppose that (AS1)-(AS3) hold and further suppose that $\{\bx^{(k)} \in B, k \in K\}$ is a sequence that converges to $\bx^*$ and $\{\hat \lambda(\bx^{(k)}) \in \R^m, k \in K\}$ is the sequence derived from  (\ref{lambdaestimate}). Then, suppose that
\begin{align} \label{supp1}
\|\bP(\bx^{(k)}, \nabla_{\bx}L_{\mu}^{(k)} ) \|_2\leq \omega^{(k)}
\end{align}
where $\omega^{(k)}$ refers to positive scalar parameters that converge to zero with an increase in $k \in K$ and $\nabla_{\bx}L_{\mu}^{(k)}$ is defined in (\ref{gradientofL}). \\
(i)  We have constant $a_2 > 0$ and integral $k_0$, such that for all $k \geq k_0$ and $k \in K$
\begin{align}  \label{mainlem1}
\|\hat \lambda(\bx^{(k)}) - \hat \lambda(\bx^*) \|_2 \leq a_2 \| \bx^{(k)} - \bx^*\|_2.
\end{align}
(ii) We also have positive constants $a_1$, $a_2$, and integer $k_0$, such that for all $k \geq k_0$, $k \in K$, 
\begin{align} \label{mainlem2}
\| \bar \lambda^{(k)} - \hat \lambda(\bx^*)\|_2 \leq a_1 \omega^{(k)} + a_2 \|\bx^{(k)} - \bx^*\|_2,
\end{align}
where $\bar \lambda^{(k)}$ is defined in (\ref{barlambda}). Sequence $\{\bar \lambda^{(k)}\}$ converges to $\hat \lambda(\bx^*)$ as $k \in K$ increases.\\
(iii) $\nabla_{\bx}  L_{\mu}^{(k)} -   \nabla_{\bx} L(\bx^*, \hat \lambda(\bx^*)) \rightarrow 0 $ as $k \in K$ increases. \\
(iv) In addition, suppose that $\{\mu^{(k)}, k \in K\}$ forms a non-increasing sequence of positive scalars, $\mu^{(k)} \rightarrow 0$, and $\mu^{(k)}\lambda^{(k)} \rightarrow 0$ as $k \in K$ increases. Then, we obtain positive constants $a_1$, $a_2$, and integer $k_0$, such that for $k \geq k_0$, $k \in K$ and for any $c_i$
{\small
\begin{align} \label{mainlem3}
\|c_i(\bx^{(k)}) \|_2 \leq \mu^{(k)} (a_1 \omega^{(k)}+ a_2 \| \bx^{(k)} - \bx^*\|_2+ \|\lambda^{(k)} - \hat \lambda(\bx^*)\|_2  ).
\end{align}
}Therefore, $c_i(\bx^*) = 0$ as $c_i$ is a continuous function.
\end{citedlem}

\proof
See Section \ref{appendixC}.

\qed

\begin{citedlem} \label{glolem3}
Suppose that the assumptions of Lemma \ref{glolem2} hold and further suppose that $c(\bx^*) = 0$. 
Then, $\bx^*$ is a critical point (holding the Karush-Kuhn-Tucker (KKT) condition) for (\ref{P11})-(\ref{P3}) and $\lambda^* = \hat \lambda(\bx^*)$ is the corresponding vector of Lagrangian multipliers.
\end{citedlem}

\proof
See Section \ref{appendixC}.
\qed

We are now ready to prove Theorem \ref{OURmaincriticalpoint}.

\proof 

The CGT-algorithm guarantees that when $k$ increases, $\{\mu^{(k)}\}$ and $\{\omega^{(k)}\}$ are non-increasing sequences of positive scalars and  $\omega^{(k)} \rightarrow 0$ and  $\mu^{(k)} \rightarrow 0$. The algorithm also guarantees that sequence $\{\mu^{(k)} \|\lambda^{(k)}\|, k \in K\}$ converges to zero as $k \in K$ increases (Lemma 4.2 in \cite{conn1991globally}); therefore, $\mu^{(k)} \lambda^{(k)} \rightarrow 0$. In accordance with Lemma \ref{glolem2}(iv) and  Lemma \ref{glolem3}, we obtain $c_i(\bx^{(k)}) \rightarrow 0$ and  $c_i(\bx^*) = 0$ and $\bx^*$ is a critical point for (\ref{P11})-(\ref{P3}).

Finally, we need to demonstrate that (\ref{supp1}) holds for any $k$, provided that the minimizers of the augmented Lagrangian function  $\nabla  L_{\mu^{(k)}}(\bx^{(k)}, \lambda^{(k)})$ are critical points with fixed $\mu^{(k)}$ and $\lambda^{(k)}$ for all $k$. 
Without a loss of the generality, we suppose that the critical point $\bx^{(k)} \in B$ is a limit point of the sequence $\{\bz^{(l)} \in B\}$ with $\bz^{(0)} = \bx^{(k-1)} \in B$. 
This means that  the sequence of the $i$-th component $\{\bz_i^{(l)}\}$ of $\{\bz^{(l)}\}$ approaches $\bx^{(k)}_i$ through 
$(\bz_i^{(l)} \geq l_i ) \rightarrow l_i$, $(\bz_i^{(l)} \leq u_i )\rightarrow u_i$, or $(\nabla L_{\mu^{(k-1)}} (\bz^{(l)}, \lambda^{(k-1)}) )_i\rightarrow 0$. In accordance with (\ref{dominatedabove}) to (\ref{floating1}), as $l$ increases,
\begin{align}
\|\bP(\bz^{(l)}, \nabla  L_{\mu^{(k-1)}} (\bz^{(l)}, \lambda^{(k-1)}))\| \rightarrow 0.
\end{align}
There exists $l_0$ (depending on $\omega^{(k-1)}$), such that $l \geq l_0$ and 
\begin{align}
\|\bP(\bz^{(l)}, \nabla  L_{\mu^{(k-1)}} (\bz^{(l)}, \lambda^{(k-1)}))\|\leq \omega^{(k-1)}.
\end{align}
This completes the proof of Theorem \ref{OURmaincriticalpoint}.

\qed

\subsection{Proofs of Lemmas \ref{glolem2} and \ref{glolem3}} \label{appendixC}

\subsubsection{Proof of Lemma \ref{glolem2}}

We show parts (iii) and  (iv). Proofs of the other parts can be found in \cite{conn1991globally}. 

(iii) 
For $k \geq k_0$, we obtain
\begin{align*}
& \|\nabla_{\bx}  L_{\mu}^{(k)} -  \nabla_{\bx} L(\bx^*, \hat \lambda(\bx^*))\|  \\
= & \|\nabla f(\bx^{(k)})+\bA(\bx^{(k)}) \bar \lambda^{(k)} - [ \nabla f(\bx^*) + \bA(\bx^*) \hat  \lambda(\bx^*)] \| \\
\leq & \|\nabla f(\bx^{(k)}) - \nabla f(\bx^*)\|+ \| \bA(\bx^*)\|  \| \bar \lambda^{(k)} - \hat \lambda(\bx^*)\| \\
+ & \|\bA(\bx^{(k)}) - \bA(\bx^*) \| \| \bar \lambda^{(k)}\| \rightarrow 0.
\end{align*}
The conclusion follows from (i) and (ii) and $\bA(\bx^{(k)}) \rightarrow \bA(\bx^*)$ when $k\in K$ is sufficiently large, due to the fact that $\bA(\bx)$ is continuous, the derivation of which is based on (AS1) for $\bx \in B$. \\

(iv) From (\ref{barlambda}), we have
\begin{align*}
 c(\bx^{(k)}) = & \mu^{(k)} (\bar \lambda^{(k)} -  \lambda^{(k)}) \nonumber \\
 = & \mu^{(k)} [\bar \lambda^{(k)} - \hat \lambda(\bx^*) + (\hat \lambda(\bx^*) -   \lambda^{(k)} ) ].
\end{align*}
Using (ii) for all $k \geq k_0$ and $k \in K$, we obtain
{\small{
\begin{align*} 
\| c(\bx^{(k)}) \| \leq \mu^{(k)}(a_1 \omega^{(k)} + a_2 \| \bx^{(k)} - \bx^*\|+ \| \lambda^{(k)}- \hat \lambda(\bx^*)\|).
\end{align*}
}}

Following the assumptions that $\omega^{(k)} \rightarrow 0$ and $\bx^{(k)} \rightarrow \bx^*$, $a_1 \omega^{(k)} + a_2 \| \bx^{(k)} - \bx^*\|\rightarrow 0$ for sufficiently large $k \in K$. (AS3) guarantees that 
$\mu^{(k)} \rightarrow 0$, $\mu^{(k)}\lambda^{(k)} \rightarrow 0$, and $\hat \lambda(\bx^*)$ is well-defined. This implies that $\mu^{(k)}(\lambda^{(k)}-\hat \lambda(\bx^*))\rightarrow 0$. Therefore, $\mu^{(k)}\| \lambda^{(k)}- \lambda(\bx^*)\| \rightarrow 0$.
Thus, we conclude that $c(\bx^{(k)}) \rightarrow 0$. In addition, $c(\bx^{(k)}) \rightarrow c(\bx^*) = 0$ as each $c_i$ is a continuous function and $\bx^{(k)} \rightarrow \bx^*$.

\subsubsection{Proof of Lemma \ref{glolem3}}

In accordance with Lemma \ref{glolem2}(ii), sequence $\{\bar \lambda^{(k)}\}$ converges to $\hat \lambda(\bx^*)$ as $k \in K$ increases. Let $\lambda^*$ denote $\hat \lambda(\bx^*)$.
We show that $\bx^*$ and $\lambda^*$ satisfy the KKT conditions by considering variables in sets $I_0- I_4$ on a case-by-case basis. In the following, the Lagrangian multipliers $\gamma_{i,l} \geq 0$ and $\gamma_{i,u} \geq 0$ are respectively associated with the constraints $(l_i - x_i) \leq 0$ and $(x_i - u_i) \leq 0 $ for variable $x_i$ in $\bx^2$.  In addition to $c(\bx^*) = 0$, other KKT conditions for $I_0 -I_4$ are derived as follows:

\noindent Case 1. $I_0$: $i \in I_0$ indicates that $x_i$ is in $\bx^1$. From (\ref{floating1}), Lemma \ref{glolem1}(ii), and Lemma \ref{glolem2}(iii), we can conclude that $\bx^*$ and $\lambda^*$ satisfy 
\begin{align*}
0 = (\nabla_{\bx} L(\bx^*, \lambda^*))_i.
\end{align*}
\noindent Case 2. $I_1$:  $i \in I_1$ indicates that $x_i$ is in $\bx^2$ where $\bx^*$ resides within $B$. The fact that  (\ref{floating2}) holds indicates that there exists $k_0$ such that all $k \geq k_0$, $k \in K$, and $l_i  < x_i^{(k)} < u_i$. The constraints $(l_i - x_i^{(k)}) \leq 0$ and $(x_i^{(k)} - u_i) \leq 0 $ are inactive; therefore, $\gamma_{i,l}^* = 0$ and $\gamma_{i,u}^* = 0$. From Lemma \ref{glolem1}(ii) and Lemma \ref{glolem2}(iii), we can conclude that $\bx^*$ and $\lambda^*$ satisfy 
\begin{align*}
0 = (\nabla_{\bx} L(\bx^*, \lambda^*))_i.
\end{align*}
\noindent Case 3. $I_{3u}$ and $I_{3l}$:  We consider the case where $x_i$ is a component of $\bx^2$ and $i \in I_{3u}$ (the case where $i \in I_{3l}$ can be derived using a similar method and is therefore omitted for brevity). In accordance with Lemma \ref{glolem1}(i), there exists $k_0$ such that all $k  \geq k_0$, $k \in K$, and $x_i^{(k)} > l_i$.  Constraint $x_i^{(k)} \geq l_i$ is inactive; therefore, $\gamma_{i,l}^* = 0$.
The Lagrangian function with respect to $x_i$ for $k \geq k_0$ is 
\begin{align}
f(\bx^{(k)}) + \sum_{i=1}^m \lambda_i c_i(\bx^{(k)}) + \sum_{i \in I_{3u}}\gamma_{i,u}^{(k)} (x_i^{(k)}-u_i).
\end{align}
We demonstrate that the following conditions are met at the limit:
\begin{align*}
\begin{cases}
0 = ( \nabla_{\bx} L(\bx^*, \lambda^*))_i + \gamma_{i,u}^*, \\
\gamma_{i,u}^*(x_i^* - u_i) = 0, \\
\gamma_{i,u}^* \geq 0.
\end{cases}
\end{align*}
The complementary slackness condition $\gamma_{i,u}^*(x_i^* - u_i) = 0$ holds, based on the definition of $i \in I_{3u}$, which implies that $x_i^* - u_i = 0$. $( \nabla_{\bx} L(\bx^*, \lambda^*))_i = 0$ and $\gamma^*_{i,u} = 0$ can be derived from (\ref{floating2}), Lemma \ref{glolem1}(ii), and Lemma \ref{glolem2}(iii). 

\noindent Case 4. $I_{2l}$: 
According to Lemma \ref{glolem1}(i) and (\ref{dominatedabove}), there exists $k_0$ such that for all $k \geq k_0$, $k \in K$, $x_i\leq u_i$ is inactive; therefore, $\gamma_{i,u}^* = 0$.  The Lagrangian associated with the variable for $k \geq k_0$ is 
\begin{align}
f(\bx^{(k)}) + \sum_{i=1}^m \lambda_i c_i(\bx^{(k)}) + \sum_{i \in I_{2l}}\gamma_{i,l}^{(k)} (l_i - x_i^{(k)}).
\end{align}
We demonstrate that the following conditions are met at the limit:
\begin{align} \label{KKT2l}
\begin{cases}
0 = (\nabla_{\bx} L(\bx^*, \lambda^*))_i - \gamma_{i,l}^*, \\
\gamma_{i,l}^*(l_i - x_i^*) = 0, \\
\gamma_{i,l}^* \geq  0.
\end{cases}
\end{align}
Complementary slackness holds, due to Lemma \ref{glolem1}(i) where $l_i - x_i^* = 0$.
Based on (\ref{case1}) and Lemma \ref{glolem2}(iii), $(\nabla_{\bx} L(\bx^*, \lambda^*))_i \geq x_i^* - l_i   = 0$. Thus, $\gamma_{i,l}^* \geq  0$ and $0 = (\nabla_{\bx} L(\bx^*, \lambda^*))_i - \gamma_{i,l}^*$. 

\noindent Case 5. $I_{2u}$: 
Based on Lemma \ref{glolem1}(i) and (\ref{dominatedbelow}), there exists $k_0$ such that for all $k \geq k_0$, $k \in K$, and $x_i^{(k)}  \geq l_i$ is inactive; therefore, $\gamma_{i,l}^* = 0$. The Lagrangian associated with the variable where $k \geq k_0$ is as follows:
\begin{align}
f(\bx^{(k)}) + \sum_{i=1}^m \lambda_i c_i(\bx^{(k)}) + \sum_{i \in I_{2u}}\gamma_{i,u}^{(k)} (x_i^{(k)} - u_i).
\end{align}
We demonstrate that the following conditions are met at the limit:
\begin{align} \label{KKT2u}
\begin{cases}
0 = (\nabla_{\bx} L(\bx^*, \lambda^*))_i + \gamma_{i,u}^*, \\
\gamma_{i,u}^*(x_i^* - u_i) = 0, \\
\gamma_{i,u}^* \geq 0.
\end{cases}
\end{align}
According to Lemma \ref{glolem1}(i) where $x_i^* - u_i =0$, complementary slackness condition holds.
Based on Lemma \ref{glolem2}(iii) and (\ref{case2}), $(\nabla_{\bx} L(\bx^*, \lambda^*))_i  \leq x_i^* - u_i = 0$ for  sufficiently large $k$. Thus, $\gamma_{i,u}^* \geq 0$ and $0 = (\nabla_{\bx} L(\bx^*, \lambda^*))_i + \gamma_{i,u}^*$. 

\noindent Case 6. $I_4$:  Let $x_i$ be a component in $\bx^2$ and $i \in I_4$. 
We consider a subsequence of $k \in K$ with variables with index $i \in I_4$ that converges to $x_i^* = l_i$ when $k \in K$ is sufficiently large. We omit the derivations when the sub-sequences converge to $u_i$ or inside $[l_i, u_i]$ because they can be dealt with in a similar manner. Constraint $x_i^* \leq u_i$ is inactive for the subsequence of $k \in K$, where $k$ is sufficiently large; therefore, $\gamma_{i,u}^* = 0$. From Lemma \ref{glolem1}(iii) and Lemma \ref{glolem2}(iii), $(\nabla_{\bx} L_{\mu}^{(k)})_i$ converges to $(\nabla_{\bx} L(\bx^*, \lambda^*))_i$ and $(\nabla_{\bx} L(\bx^*, \lambda^*))_i = 0$ for $k \in K$ and $k$ is sufficiently large.
Setting $\gamma_{i,l}^* = 0$ meets the following conditions at the limit :\begin{align*}
\begin{cases}
0 =  (\nabla_{\bx} L(\bx^*, \lambda^*))_i - \gamma_{i,l}^* \\
\gamma_{i,l}^*(l_i - x_i^*) = 0, \\
\gamma_{i,l}^* \geq 0.
\end{cases}
\end{align*}
\qed
\section{Primal variable updating of (P2)} \label{appendixprimalvariable}

The augmented Lagrangian  function of (P2) exclusively involves equality constraints (denoted as $\mathcal L_{2,\bar \theta}$),  and can be expressed as follows:
\begin{align} \label{2aug00}
& \frac{1}{2}\sum_{j}\|\by_j - M_2  \bv_j\|^2 + \sum_{ji} [\frac{\rho_1}{2}\|v_{ji} -d_{ji} u_{ji}\|^2 +\frac{\rho_2}{2}\|u_{ji} -(M_1 \bx_j)_i\|^2 + \frac{\rho_3}{2} \| d_{ji} u_{ji} - s_{ji} \|^2 \nonumber\\
& + \frac{\rho_4}{2} \|(1-d_{ji}) u_{ji} + t_{ji} \|^2] +\frac{c_1}{2}(\|\bW_1\|_F^2 + \|\bW_2\|_F^2) + \frac{c_2}{2} \sum_j \|\bD_j\|_F^2 + \sum_{ji} [ \bar \mu_{1ji} (v_{ji} -d_{ji} u_{ji}) \nonumber \\
& + \bar \mu_{2ji} (u_{ji} -(M_1 \bx_j)_i) + \bar \mu_{3ji} (d_{ji} u_{ji} - s_{ji}) + \bar \mu_{4ji} ((1-d_{ji}) u_{ji} + t_{ji})],
\end{align}
where $\rho_1$, $\rho_2$, $\rho_3$, $\rho_4 > 0$ are parameters, 
$\{\bar \mu_{1ji} \in \R\}_{j,i}$, $\{\bar \mu_{2ji} \in \R\}_{j,i}$, $\{\bar \mu_{3ji} \in \R\}_{j,i}$, $\{\bar \mu_{4ji} \in \R\}_{j,i}$ are Lagrangian multipliers, and $\bar \theta = \{c_1, c_2, \rho_1, \rho_2, \rho_3, \rho_4\}$.
To simplify notation, we let $\bar d = \{d_{ji}\}$, $\bar u = \{u_{ji}\}$, $\bar \bv = \{\bv_{j}\}$, $\bar s= \{s_{ji}\}$,  $\bar t= \{t_{ji}\}$, and $\bar\mu  = \{\bar \mu_{1ji}, \bar \mu_{2ji}, \bar \mu_{3ji},\bar \mu_{4ji}\}$.

The primal variables of $\mathcal L_{2,\bar \theta}$, in which dual variables $\bar\mu$ are fixed, can  alternatively be updated in reverse order as presented in Algorithm \ref{alt:Uzawa200}, where orthogonal projections to $[0, 1]$ and $\R_+$ are applied to respective variables to ensure values of $d_{j,i} \in [0, 1]$ and $t_{ji}, s_{ji} \geq 0$. 
{\tiny{
\begin{algorithm}[htb]
\caption{Primal variable update of (P2) when dual variables are fixed}
\label{alt:Uzawa200}
\algsetup{indent=2em}
\begin{algorithmic}[1]
\STATE Primal variables in $\mathcal M_2$ following the alternating optimization method are updated until the condition required by the CGT-algorithm is reached. 
\STATE INPUT:  dual variables $\{\bar \mu^{(l)} \}$. 
\STATE $k = 0$;
\REPEAT
\STATE $\bW_2^{(k+1)}   =  \arg\min_{\bW_2}\mathcal L_{2, \bar \theta}(\bW_2, \bb_2^{(k)}, \bar \bv^{(k)}, \bar d^{(k)}, \bar u^{(k)}, \bar s^{(k)}, \bar t^{(k)}, \bW_1^{(k)}, \bb_1^{(k)}; \bar\mu^{(l)})$
\STATE $\bb_2^{(k+1)}   = \arg\min_{\bb_2} \mathcal L_{2, \bar \theta}(\bW_2^{(k+1)}, \bb_2, \bar \bv^{(k)}, \bar d^{(k)} , \bar u^{(k)}, \bar s^{(k)}, \bar t^{(k)},\bW_1^{(k)}, \bb_1^{(k)}; \bar \mu^{(l)})$
\FOR{$j= 1$ to $N$} 
\STATE $\bv_{j}^{(k+1)}    = \arg \min_{\bv_{j}\geq 0} \mathcal L_{2, \bar \theta}(\bW_2^{(k+1)}, \bb_2^{(k+1)},\bv_{j},\bar d^{(k)},\bar u^{(k)}, \bar s^{(k)}, \bar t^{(k)},\bW_1^{(k)},  \bb_1^{(k)}; \bar\mu^{(l)})$
\ENDFOR
\FOR{$j= 1$ to $N$} 
\FOR{$i=1$ to $N_0$ with $i_0 < i < i_1$}
\STATE $d_{ji}^{(k+1)}    = \arg \min_{d_{ji}\in [0, 1]} \mathcal L_{2,\bar \theta}(\bW_2^{(k+1)}, \bb_2^{(k+1)}, \bar \bv^{(k+1)}, \{d_{ji_0}^{(k+1)}\},d_{ji},\{d_{ji_1}^{(k)}\},\bar u^{(k)},\bar s^{(k)}, \bar t^{(k)}, \bW_1^{(k)},  \bb_1^{(k)}; \bar\mu^{(l)})$
\ENDFOR
\ENDFOR
\FOR{$j= 1$ to $N$} 
\FOR{$i=1$ to $N_0$ with $i_0 < i < i_1$}
\STATE $u_{ji}^{(k+1)}  = \arg \min_{u_{ji}} \mathcal L_{2,\bar \theta}(\bW_2^{(k+1)}, \bb_2^{(k+1)}, \bar \bv^{(k+1)}, \bar d^{(k+1)}, \{u_{ji_0}^{(k+1)}\},u_{ji},\{u_{ji_1}^{k}\}, \bar s^{(k)}, \bar t^{(k)}, \bW_1^{(k)},  \bb_1^{(k)}; \bar \mu^{(l)})$
\ENDFOR
\ENDFOR
\FOR{$j= 1$ to $N$} 
\FOR{$i=1$ to $N_0$ with $i_0 < i < i_1$}
\STATE $s_{ji}^{(k+1)}  = \arg \min_{s_{ji}\geq 0} \mathcal L_{2,\bar \theta}(\bW_2^{(k+1)}, \bb_2^{(k+1)}, \bar \bv^{(k+1)}, \bar d^{(k+1)}, \bar u^{(k)}, \{s_{ji_0}^{(k+1)}\},s_{ji},\{s_{ji_1}^{(k)}\}, \bar t^{(k)}, \bW_1^{(k)},  \bb_1^{(k)}; \bar \mu^{(l)})$
\ENDFOR
\ENDFOR
\FOR{$j= 1$ to $N$} 
\FOR{$i=1$ to $N_0$ with $i_0 < i < i_1$}
\STATE $t_{ji}^{(k+1)}  = \arg \min_{t_{ji}\geq 0} \mathcal L_{2, \bar \theta}(\bW_2^{(k+1)}, \bb_2^{(k+1)}, \bar \bv^{(k+1)}, \bar d^{(k+1)}, \bar u^{k}, \bar s^{k}, \{t_{ji_0}^{(k+1)}\},t_{ji},\{t_{ji_1}^{(k)}\}, \bW_1^{(k)},  \bb_1^{(k)}; \bar \mu^{(l)})$
\ENDFOR
\ENDFOR
\STATE $\bW_1^{(k+1)}   = \arg\min_{\bW_1} \mathcal L_{2,\bar \theta}(\bW_2^{(k+1)},\bb_2^{(k+1)},\bar \bv^{(k+1)}, \bar d^{(k+1)}, \bar u^{(k+1)}, \bar s^{(k)}, \bar t^{(k)},\bW_1,  \bb_1^{(k)}; \bar \mu^{(l)})$
\STATE $\bb_1^{(k+1)}   = \arg\min_{\bb_1} \mathcal L_{2,\bar \theta}(\bW_2^{(k+1)}, \bb_2^{(k+1)},\bar \bv^{(k+1)}, \bar d^{(k+1)}, \bar u^{(k+1)}, \bar s^{(k)}, \bar t^{(k)},\bW_1^{(k+1)}, \bb_1; \bar \mu^{(l)})$
\STATE $k = k+1;$
\UNTIL certain stop condition required by the CGT-algorithm is met.
 \RETURN 
\end{algorithmic}
\end{algorithm}
}}

The unique minimizer in updating $\bW_1$, $\bb_1$, $\bW_2$, $\bb_2$, $\{d_{ji}\}$, $\{u_{ji}\}$, $\{v_{ji}\}$, $\{s_{ji}\}$, and $\{t_{ji}\}$ of Algorithm \ref{alt:Uzawa200}can be efficiently derived, due to the fact that the augmented Lagrangian function is a strongly quadratic convex function when viewed as a function of any of the above variables with the other variables fixed. For example, using the gradient descent method, ${\cal O}(\ln \frac{1}{\epsilon})$ is required to reach the $\epsilon$-suboptimal solution. In the following, we present closed-form solutions for primal variable updates, where $\bu_j = [ u_{ji}]_i$, $\bv_j = [ v_{ji}]_i$, $\bs_j = [ s_{ji}]_i$, and $\bt_j = [t_{ji}]_i$ are vectors of slack variables associated with data $\bx_j$. $\bar \mu_{1j} = [ \bar \mu_{1ji}]_i$, $\bar \mu_{2j} = [ \bar \mu_{2ji}]_i$, $\bar \mu_{3j} = [ \bar \mu_{3ji}]_i$, and $\bar \mu_{4j} = [ \bar \mu_{4ji}]_i$ are Lagrangian multipliers associated with $\bx_j$, and $N$ indicates the size of the training data. Note that the most recently variable estimates are used in place of the following updates, as demonstrated in Algorithm \ref{alt:Uzawa200}.

\noindent{$\bullet$} Updating $\bW_1$ and $\bW_2$, we obtain the following:
\begin{align*}
 \bW_1 & = (\sum_j \rho_2(\bu_j-\bb_1) \bx_j^\top+ \bar \mu_{2j})(\sum_j \rho_2 \bx_j \bx_j^\top + c_1 \bI)^{-1},\\
 \bW_2 & = (\sum_j (\by_j-\bb_2) \bv_j^\top)(\sum_j \bv_j \bv_j^\top + c_1 \bI)^{-1},
 \end{align*}
where $c_1 > 0$.

\noindent{$\bullet$} Updating $\bb_1$ and $\bb_2$, we obtain the following:
\begin{align*}
 \bb_1& = \frac{1}{N}(\sum_j \bu_j - \bW_1 \bx_j+\frac{1}{\rho_2}\bar \mu_{2j}),\\
 \bb_2 & =\frac{1}{N}(\sum_j \by_j - \bW_2 \bv_j).
 \end{align*}

\noindent{$\bullet$} Updating $d_{ji}$, we obtain the following:
\begin{align*}
d_{ji} = \bP_{[0,1]}\big[\frac{(\rho_1 v_{ji} + \rho_3 s_{ji} + \rho_4( u_{ji}+t_{ji}) + \bar\mu_{1ji}-\bar\mu_{3ji}+\bar\mu_{4ji})u_{ji}}{ (1+\rho_3+\rho_4)u_{ji}^{2} + c_2}\big],
 \end{align*}
where $\bP_{[0,1]}$ denotes the orthogonal projection of a scalar to the interval $[0, 1]$.  Note that $c_2$ in denominator avoids dividing by a zero in updating $d_{ji}$.

\noindent{$\bullet$} Updating $\bu_{j}$, we obtain the following:
\begin{align*}
\bu_{j} & =(\rho_1 \bD_{j}^\top \bD_{j} +\rho_2 \bI + \rho_3 \bD_{j}^\top \bD_{j} +\rho_4 (\bI-\bD_{j})^\top (\bI-\bD_{j}))^{-1} (\rho_1 \bD_{j}^\top \bv_{j} + \rho_2 (\bW_1 \bx_j +\bb_1) \\
&+ \rho_3 \bD_{j}^\top \bs_{j} + \rho_4( \bD_{j}-\bI)^\top \bt_{j} + \bD_{j}^\top \bar\mu_{1j} - \bar\mu_{2j}-\bD_{j}^\top \bar\mu_{3j}-(\bI-\bD_j)^\top \bar\mu_{4j}).
 \end{align*}

\noindent{$\bullet$} Updating $\bv_{j}$, we obtain the following:
\begin{align*}
\bv_{j} = (\bW_2^\top \bW_2 +\rho_1 \bI)^{-1}(\bW_2^\top (\by_j - \bb_2)+\rho_1 \bD_j \bu_j-\bar \mu_{1j}).
 \end{align*}

\noindent{$\bullet$} Updating vectors $\bs_j$ and $\bt_j$, we obtain the following:
\begin{equation}
\begin{aligned}
\bs_{j} &=\bP_{\geq 0} (\bD_{j} \bu_j + \frac{\bar{\mu}_{3j}}{\rho_3}),\\
\bt_{j} &=\bP_{\geq 0}((\bD_{j}-\bI) \bu_j - \frac{\bar{\mu}_{4j}}{\rho_4}),\\
\end{aligned}
\end{equation}
where $\bP_{\geq 0}$ is a point-wise operation which denotes the orthogonal projection of each element of a vector to a non-negative value.

\section{Primal variable updating of (PL)} \label{appD}

Note that in the following, $j=1, \cdots N$; $k=1, \cdots, L-1$; $i=1, \cdots, N_k$; and $\bv_j^0 = \bx_j$. The augmented Lagrangian  function of (PL) involving the equality constraints is denoted as $\mathcal L_{L,\bar \theta}$ and can be expressed as follows:
\begin{align} \label{2aug00}
& \frac{1}{2}\sum_{j}\|\by_j - M_L  \bv_j^{L-1}\|^2 + \sum_{jik} [\frac{\rho_1}{2}\|v^k_{ji} -d^k_{ji} u^k_{ji}\|^2 +\frac{\rho_2}{2}\|u^k_{ji} -(M_k \bv^{k-1}_j)_i\|^2 + \frac{\rho_3}{2} \| d^k_{ji} u^k_{ji} - s^k_{ji} \|^2 \nonumber\\
& + \frac{\rho_4}{2} \|(1-d^k_{ji}) u^k_{ji} + t^k_{ji} \|^2] +\frac{c_1}{2}\sum_{l=1}^L\|\bW_l\|_F^2  + \frac{c_2}{2} \sum_{jk}\|\bD^k_j\|_F^2 + \sum_{jik} [ \mu^k_{1ji} (v^k_{ji} -d^k_{ji} u^k_{ji}) \nonumber \\
& + \mu^k_{2ji} (u^k_{ji} -(M_k \bv^{k-1}_j)_i) + \mu^k_{3ji} (d^k_{ji} u^k_{ji} - s^k_{ji}) + \mu^k_{4ji} ((1-d^k_{ji}) u^k_{ji} + t^k_{ji})],
\end{align}
where $\rho_1$, $\rho_2$, $\rho_3$, $\rho_4 > 0$ are parameters, 
$\{\mu^k_{1ji} \in \R\}_{j,i,k}$, $\{\mu^k_{2ji} \in \R\}_{j,i,k}$, $\{\mu^k_{3ji} \in \R\}_{j,i,k}$, $\{\mu^k_{4ji} \in \R\}_{j,i,k}$ are Lagrangian multipliers, and $\bar \theta = \{c_1, c_2, \rho_1, \rho_2, \rho_3, \rho_4\}$.
To simplify notation, we let $\bar d^k = \{d^k_{ji}\}$, $\bar u^k = \{u^k_{ji}\}$, $\bar \bv^k = \{\bv^k_{j}\}$, $\bar s^k= \{s^k_{ji}\}$,  $\bar t^k= \{t^k_{ji}\}$, and $\bar\mu  = \{\bar \mu^k_{1ji}, \bar \mu^k_{2ji}, \bar \mu^k_{3ji},\bar \mu^k_{4ji}\}$ where $\bar \mu^k_{1j} = \{\mu^k_{1ji}\}_{i}$, $\bar \mu^k_{2j} = \{\mu^k_{2ji}\}_{i}$, $\bar \mu^k_{3j} = \{\mu^k_{3ji}\}_{i}$, and $\bar \mu^k_{4j} = \{\mu^k_{4ji}\}_{i}$.

The primal variables of $\mathcal L_{L,\bar \theta}$ (where dual variables $\bar\mu$ are fixed) can  alternatively be updated in reverse order from layer $L$ to layer $1$ as follows in which the orthogonal projections to $[0, 1]$ and $\R_+$ are applied to respective variables to ensure that $d^k_{j,i} \in [0, 1]$ and $t^k_{ji}, s^k_{ji} \geq 0$.  Algorithm \ref{funrec}
outlines the procedure used in the updating of primal variables of (PL) when dual variables are fixed.

\begin{algorithm}[htb]
\caption{Primal variable update of (PL) when dual variables are fixed}
\label{funrec}
\algsetup{indent=2em}
\begin{algorithmic}[1]
\STATE Primal variables in $\mathcal M_L$ following the alternating optimization method are updated when dual variables are fixed.
\STATE  INPUT: dual variables and the iteration index, $k$.
\STATE Update $\bW_L^{(k-1)} \rightarrow \bW_L^{(k)}$ by solving (\ref{AlderW}).
\STATE Update $\bb_L^{(k-1)} \rightarrow \bb_L^{(k)}$ by solving (\ref{Alderb}).
\FOR{$l= L-1$ to $1$} 
\STATE Update $\{(\bv^{l}_{j})^{(k-1)}\}_{j=1}^N \rightarrow \{(\bv^{l}_{j})^{(k)}\}_{j=1}^N$ by solving  (\ref{Alderv}).
\STATE Update $\{\{(d^{l}_{ji})^{(k-1)}\}_{i=1}^{N_{l}}\}_{j=1}^N \rightarrow \{\{(d^{l}_{ji})^{(k)}\}_{i=1}^{N_{l}}\}_{j=1}^N$ by solving  (\ref{Alderd}).
\STATE  Update $\{(\bu^{l}_{j})^{(k-1)}\}_{j=1}^N \rightarrow \{(\bu^{l}_{j})^{(k)}\}_{j=1}^N$ by solving (\ref{Alderu}).
\STATE  Update $\{(\bs^{l}_{j})^{(k-1)}\}_{j=1}^N \rightarrow \{(\bs^{l}_{j})^{(k)}\}_{j=1}^N $ by solving (\ref{Alderst}).
\STATE Update $\{(\bt^{l}_{j})^{(k-1)}\}_{j=1}^N \rightarrow \{(\bt^{l}_{j})^{(k)}\}_{j=1}^N $ by solving (\ref{Alderst}).
\STATE Update $\bW_l^{(k-1)} \rightarrow \bW_l^{(k)}$ by solving (\ref{AlderW}).
\STATE Update $\bb_l^{(k-1)} \rightarrow \bb_l^{(k)}$ by solving (\ref{Alderb}).
\ENDFOR
 \RETURN 
\end{algorithmic}
\vspace{2pt} {\hrule height0.5pt} \vspace{6.5pt}
\noindent \textbf{Remark.} Many efficient methods have been devised to derive unique solutions to updates (\ref{AlderW})-(\ref{Alderst}), due to the fact that they are strongly-convex problems, involving quadratic minimization. Note that the order of variables involved in updating must strictly follow the order of the alternating minimization method, an example of which is detailed in  Algorithm \ref{alt:Uzawa200}. 
\end{algorithm}

Note that the following updating of variables must proceed in the order of the alternating minimization method, where the most recently estimated variables are adopted for the following update.

\noindent{$\bullet$} $\bW_l$ and $\bW_L$ (where $l = 1, \cdots, L-1$):  
\begin{align} \label{AlderW}
\begin{cases}
\min_{\bW_l} \sum_{j=1}^N \{\frac{\rho_2}{2} \|\bu_j^{l} - \bW_l \bv_j^{l-1}-\bb_{l}\|^2\ +(\bar \mu_{2j}^{l})^\top(\bu_j^{l} - \bW_l \bv_j^{l-1}-\bb_{l}) \} +  \frac{c_1}{2} \|\bW_l\|_F^2,\\
\min_{\bW_L} \sum_{j=1}^N \{\frac{1}{2} \|\by_j - \bW_L \bv_j^{L-1}-\bb_{L}\|^2 \} +  \frac{c_1}{2} \|\bW_L\|_F^2.  
\end{cases}
\end{align}
where $\bv_j^{0}=\bx_j$.

\noindent{$\bullet$} $\bb_l$ and $\bb_L$ (where $l = 1, \cdots, L-1$):
\begin{align}\label{Alderb}
\begin{cases}
\min_{\bb_l} \sum_{j=1}^N \{\frac{\rho_2}{2} \|\bu_j^{l} - \bW_l \bv_j^{l-1}-\bb_{l}\|^2 + (\bar \mu_{2j}^{l})^\top(\bu_j^{l} - \bW_l \bv_j^{l-1}-\bb_{l}) \},\\
\min_{\bb_L} \sum_{j=1}^N \{\frac{1}{2} \|\by_j - \bW_L \bv_j^{L-1}-\bb_{L}\|^2 \}.
\end{cases}
\end{align}

\noindent{$\bullet$} $d^l_{ji}$:
\begin{align} \label{Alderd}
&\min_{d^l_{ji} \in [0,1]} \frac{\rho_1}{2}\|v_{ji}^{l} -d_{ji}^{l} u_{ji}^{l}\|^2+\frac{\rho_3}{2} \| d_{ji}^{l} u_{ji}^{l} - s_{ji}^{l} \|^2+\frac{\rho_4}{2} \| (1-d_{ji}^{l}) u_{ji}^{l} + t_{ji}^{l} \|^2 \nonumber\\
& +\mu_{1ji}^{l} (v_{ji}^{l} -d_{ji}^{l} u_{ji}^{l}) +\mu_{3ji}^{l} (d_{ji}^{l} u_{ji}^{l} - s_{ji}^{l})+\mu_{4ji}^{l} ((1-d_{ji}^{l}) u_{ji}^{l} + t_{ji}^{l})+\frac{c_2}{2}(d^l_{ji})^2.
\end{align}

\noindent{$\bullet$} $\bu^l_{j}$: 
\begin{align} \label{Alderu}
&\min_{\bu_j} \frac{\rho_1}{2}\|\bv_{j}^{l} -\bD_{j}^{l} \bu_{j}^{l}\|^2 +\frac{\rho_2}{2}\|\bu_{j}^{l} -\bW_l \bv_j^{l-1} -\bb_l\|^2 + \frac{\rho_3}{2} \| \bD_{j}^{l} \bu_{j}^{l} - \bs_{j}^{l}) \|^2\nonumber\\
&+ \frac{\rho_4}{2} \| (\bI-\bD_{j}^{l}) \bu_{j}^{l} + \bt_{j}^{l}) \|^2 +(\bar \mu_{1j}^{l})^\top (\bv_{j}^{l} -\bD_{j}^{l} \bu_{j}^{l}) + (\bar \mu_{2j})^\top (\bu_{j}^{l} -\bW_l \bv_j^{l-1} -\bb_l)\nonumber\\
& +(\bar \mu_{3j})^\top (\bD_{j}^{l} \bu_{j}^{l} - \bs_{j}^{l})+(\bar \mu_{4j})^\top ((\bI-\bD_{j}^{l}) \bu_{j}^{l} + \bt_{j}^{l}). 
\end{align}

\noindent{$\bullet$} $\bv^l_{j}$:
\begin{align} \label{Alderv}
&\min_{\bv_j} \frac{\rho_1}{2}\|\bv_{j}^{l} -\bD_{j}^{l} \bu_{j}^{l}\|^2 +\frac{\rho_2}{2}\|\bu_{j}^{l+1} -\bW_{l+1} \bv_j^{l} -\bb_{l+1}\|^2 \nonumber\\
&+(\bar \mu_{1j}^{l})^\top (\bv_{j}^{l} -\bD_{j}^{l} \bu_{j}^{l}) + (\bar \mu_{2j}^{l+1})^\top (\bu_{j}^{l+1} -\bW_{l+1} \bv_j^{l} -\bb_{l+1}).
\end{align}

\noindent{$\bullet$} $\bs_j^l$ and $\bt_j^l$: 
\begin{equation} \label{Alderst}
\begin{aligned}
& \min_{\bs_j^l} \frac{\rho_3}{2}|| \bD_j^l \bu_j^{l} - \bs_j^l ||^2-(\bar{\mu}_{3j}^l)^\top (\bD_j^l \bu_j^{l} - \bs_j^l), \\
& \min_{\bt_j^l} \frac{\rho_4}{2}|| (\bI-\bD_j^l)\bu_j^{l} + \bt_j^l ||^2+(\bar{\mu}_{4j}^l)^\top ((\bI-\bD_j^l)\bu_j^{l} + \bt_j^l).
\end{aligned}
\end{equation}

If an update from (\ref{AlderW0})-(\ref{Alderst}) is performed using the gradient descent method, then ${\cal O}(\ln\frac{1}{\epsilon})$ iterations are required to achieve the $\epsilon$-suboptimal solution. 
As an alternative to the gradient descent method, closed-form unique solutions can be obtained. Let $\bar N = \max\{N_0, \cdots, N_L\}$ and $N$ denote the size of input data. In the following analysis, the inverse of an $\bar N \times \bar N$ matrix costs $\mathcal O(\bar N^3)$, the matrix-matrix multiplication of two $\bar N \times \bar N$ matrices costs $\mathcal O(\bar N^3)$, and the matrix-vector multiplication costs $\mathcal O(\bar N^2)$. 
The closed-form solutions are presented below for reference.

\noindent{$\bullet$} Updating $\mathbf{W}_l$ and $\mathbf{W}_L$:
\begin{equation} \label{AlderW0}
\begin{aligned}
& \bW_l = (\sum_j \rho_2(\bu^l_j-\bb_l) (\bv^{l-1}_j)^\top+ \bar \mu^l_{2j} (\bv^{l-1}_j)^\top)(\sum_j \rho_2 \bv^{l-1}_j (\bv^{l-1}_j)^\top + c_1 \bI)^{-1},\\
& \bW_L = (\sum_j (\by_j-\bb_L) (\bv^{l-1}_j)^\top)(\sum_j \bv^{l-1}_j (\bv^{l-1}_j)^\top + c_1 \bI)^{-1}.
\end{aligned}
\end{equation}
The degree of complexity is dominated by calculation of $\bW_l$, the complexity of which is dominated by obtaining and inverting the matrix of the second term on the right-hand side of $\bW_l$. Obtaining the matrix is on the order of $N \bar N^3$ and obtaining its inverse is $\bar N^3$. The complexity of calculating $\bW_l$ is $\mathcal O(N \bar N^3)$. The complexity of calculating all $\bW_l$ is therefore $\mathcal O(L N \bar N^3)$.

\noindent{$\bullet$} Updating $\mathbf{b}_l$ and $\mathbf{b}_L$:
\begin{equation}\label{Alderb0}
\begin{aligned}
& \bb_l = \frac{1}{N}(\sum_j \bu^l_j - \bW_l \bv^{l-1}_j+\frac{1}{\rho_2}\bar \mu^l_{2j}),\\
& \bb_L = \frac{1}{N}(\sum_j \by^l_j - \bW_L \bv^{L-1}_j).
\end{aligned}
\end{equation} 
The complexity of updating $\bb_l$ is dominated by calculation of the right-hand side of $\bb_l$, which involves matrix-vector multiplication of an $\bar N \times  \bar N$ matrix. The complexity involved in obtaining the vector is on the order of $N \bar N^2$. The complexity of obtaining $\bb_l$ is $\mathcal O(N \bar N^2)$. The complexity of obtaining all $\bb_l$ is therefore $\mathcal O(L N \bar N^2)$. 

\noindent{$\bullet$} Updating ${d}_{ji}^l$:
\begin{equation} \label{Alderd0}
\begin{aligned}
d_{ji}^{l} = P_{[0,1]}\big(\frac{(\rho_1 v_{ji}^{l} + \rho_3 s_{ji}^{l} + \rho_4( u_{ji}^{l}+t_{ji}^{l}) + \bar\mu_{1ji}^{l}-\bar\mu_{3ji}^{l}+\bar\mu_{4ji}^{l})u_{ji}^{l}}{ (1+\rho_3+\rho_4)(u_{ji}^{l})^2 + c_2}\big),
 \end{aligned}
 \end{equation}
where $P_{[0,1]}$ denotes the orthogonal projection of a scalar to interval $[0, 1]$. 
The complexity involved in obtaining $d^l_{ji}$ is on the order of $\mathcal O( \bar N)$. The complexity of obtaining all $d^l_{ji}$ is therefore $\mathcal O(L N \bar N^2)$.

\noindent{$\bullet$} Updating $\bu_{j}^l$:
\begin{equation} \label{Alderu0}
\begin{aligned}
\bu_{j}^l & =(\rho_1 (\bD_{j}^l)^\top \bD_{j}^l +\rho_2 \bI + \rho_3 (\bD_{j}^L)^\top \bD_{j}^l +\rho_4 (\bI-\bD_{j}^l)^\top (\bI-\bD_{j}^l))^{-1} (\rho_1 (\bD_{j}^l)^\top \bv_{j}^l \\
& + \rho_2 (\bW_{l} \bv_j^{l-1} +\bb_{l}) + \rho_3 (\bD_{j}^l)^\top \bs_{j}^l + \rho_4( \bD_{j}^l-\bI)^\top \bt_{j}^l + (\bD_{j}^l)^\top \bar\mu_{1j}^l - \bar\mu_{2j}^l-(\bD_{j}^l)^\top \bar\mu_{3j}^l\\
& -(\bI-\bD_j^l)^\top \bar\mu_{4j}^l),
 \end{aligned}
\end{equation}
where $\bv_j^{0}=\bx_j$.

The complexity of updating $\bu^l_j$ is dominated by obtaining and inverting the matrix of the first term on the right-hand side of $\bu^l_j$. The complexity of obtaining the matrix is on the order of $N \bar N^3$ and the complexity of obtaining its inverse is $\bar N^3$. The complexity of calculating $\bv^l_j$ is $\mathcal O(N \bar N^3)$. The complexity of calculating all $\bv^l_j$ is therefore $\mathcal O(L N \bar N^3)$.

\noindent{$\bullet$} Updating $\bv_{j}^l$ and $\bv_{j}^{L-1}$:
\begin{equation} \label{Alderv0}
\begin{aligned}
& \bv_{j}^l = \bP_{[0, \infty)} ((\rho_2 \bW_{l+1}^\top \bW_{l+1} +\rho_1 \bI)^{-1}(\rho_2\bW_{l+1}^\top (\bu_j^{l+1} - \bb_{l+1})+\rho_1 \bD_j^l \bu_j^l-\bar \mu_{1j}^l+\bW_{l+1}^\top \bar \mu_{2j}^l)),\\
& \bv_{j}^{L-1} = \bP_{[0, \infty)} ((\bW_{L}^\top \bW_{L} +\rho_1 \bI)^{-1}(\bW_{L}^\top (\by_j - \bb_{L})+\rho_1 \bD_j^{L-1} \bu_j^{L-1}-\bar \mu_{1j}^{L-1})),
 \end{aligned}
\end{equation}
where $\bP_{[0,\infty]}$ is a point-wise operation denoting the orthogonal projection of each element of a vector to interval $[0, \infty)$.
The complexity of updating $\bv^l_j$ is dominated by obtaining and inverting the matrix of the first term on the right-hand side of $\bv^l_j$. The complexity of obtaining the matrix is on the order of $N \bar N^3$ and the complexity of obtaining its inverse is $\bar N^3$. The complexity of calculating $\bv^l_j$ is $\mathcal O(N \bar N^3)$. The complexity of calculating all $\bv^l_j$ is therefore $\mathcal O(L N \bar N^3)$. 

\noindent{$\bullet$} Updating $\bs_{j}^l$ and $\bt_{j}^l$:
\begin{equation} \label{Alderst0}
\begin{aligned}
\bs_{j}^l &=\mathbf{P}_{[0,\infty]} (\bD_j^l \bu_{j}^l + \frac{\bar \mu_{3j}^l}{\rho_3}),\\
\bt_{j}^l &=\mathbf{P}_{[0,\infty]} ((\mathbf{D}_{j}^l-\mathbf{I})\bu_j^{l} - \frac{\bar \mu_{4j}^l}{\rho_4}).\\
\end{aligned}
\end{equation}

The complexity of updating $\bs_j^l$ at layer $l$ is $\mathcal O(N \bar N)$ and the complexity of updating $\bt_j^l$ is $\mathcal O(N \bar N^2)$. The complexity of obtaining all $\bs_j^l$ and $\bt_j^l$ is  therefore $\mathcal O(L N \bar N)$ and $\mathcal O(L N \bar N^2)$, respectively.

\end{document}